\documentclass{article}
\pdfoutput=1

\usepackage{arxiv}

\usepackage[utf8]{inputenc}
\usepackage[T1]{fontenc}
\usepackage[numbers,sort&compress]{natbib}
\usepackage{hyperref}
\usepackage{url}
\usepackage{booktabs}
\usepackage{amsfonts}
\usepackage{amsmath}
\usepackage{nicefrac}
\usepackage{microtype}
\usepackage{graphicx}
\usepackage{multirow}
\usepackage{array}
\usepackage{xcolor}
\usepackage[capitalize]{cleveref}

\graphicspath{{figures/}}

\newcommand{\tbf}[1]{\textbf{#1}}

\title{When Should a Network Emit Geometry, and When Should It Detect It?\\
Readout, Reconciliation, and Representation in Floorplan Vectorization}

\author{%
  \textbf{He Zhang} \\
  Independent Researcher \\
  \texttt{cyprinus12138@gmail.com} \\
}

\renewcommand{\shorttitle}{When Should a Network Emit Geometry, and When Should It Detect It?}

\hypersetup{
  pdftitle={When Should a Network Emit Geometry, and When Should It Detect It? Readout, Reconciliation, and Representation in Floorplan Vectorization},
  pdfauthor={He Zhang},
  pdfkeywords={floorplan vectorization, structured prediction, autoregressive decoding, detection, edit cost, benchmark},
}

\begin{document}
\maketitle

\begin{abstract}
A network trained to recover the walls, openings, and rooms of a rasterized
floorplan can produce its output in two ways: by emitting the geometry as an
autoregressive coordinate sequence, or by detecting it on dense junction and
centerline heatmaps and assembling a graph. We compare the two readouts on the
same trained network. On real scans (CubiCasa5K) detection is better on every
wall measure ($+2.7$ wall F1 at tolerance $0.05$, $+5.1$ at $0.015$; paired
bootstrap intervals exclude zero), and reading an opening heatmap the decoder
never used raises opening F1 by $2.6\times$ without retraining. Within real
scans the readout's advantage grows with plan size and reverses on small
plans; on clean vector renders sequence decoding is better by 5 to 8 points
where its training covered the render style, while under full domain shift
the readout, given calibrated thresholds, stays ahead; neither ink density
nor plan size explains the reversal. With matched data and recipe, a room-centric system
with a reconciliation step and a wall-first sequence model reach comparable
wall quality, so the output representation matters less than is usually
assumed. A prior from the other family helps at the output but not at the
input: deterministic fusion of the two outputs raises wall F1 by $7$ points,
whereas conditioning one model on the other's output gives no gain in three
forms, including two ground-truth-content controls. We also provide an
edit-cost metric that scores a draft by the human work needed to correct it,
corrected CubiCasa5K annotations, and ResPlan-FP, a CC~BY~4.0 benchmark of
16{,}998 plans with frozen splits and three baseline tracks. Code, the
benchmark, and the corrected annotations are available at
\url{https://github.com/Cyprinus12138/fpvec-lab}.
\end{abstract}

% =====================================================================
\section{Introduction}
% =====================================================================
Converting a rasterized floorplan into an editable vector model, consisting of
walls, openings, and labeled rooms, is a prerequisite for CAD workflows, indoor
navigation, real-estate digitization, and design tools
\citep{liu2017raster,kalervo2019cubicasa,zeng2019deepfloorplan}. In practice the
output is a draft that a person then corrects in an editor. The quantity that
matters downstream is therefore the cost of that correction rather than the F1
score of the draft, and existing IoU and F1 benchmarks do not measure it.

Once a network has been trained on this task, there are two ways to get the
geometry out of it. It can emit the walls as an autoregressive sequence of
coordinates, as sequence-based vectorizers do
\citep{phung2026raster2seq,floorplanvlm2026,hu2024rastertograph}, or it can
detect them: take the peaks of a junction heatmap as nodes, decide edges from
a centerline heatmap, and assemble a graph. The two readouts can be compared on
one and the same trained network, which removes the usual confounds of
architecture and training data. We do this for a wall-first network that has
both a sequence decoder and dense heads. On real scans (CubiCasa5K) the graph
readout is better on every wall measure, by $2.7$ points of wall F1 at
tolerance $0.05$ and $5.1$ at $0.015$ (paired bootstrap 95\% intervals
$[1.4,4.0]$ and $[3.7,6.4]$), with the largest gain at the strict tolerance
where coordinate precision matters most. On clean vector renders
(ResPlan, and a synthetic open-plan tier) the order reverses and sequence
decoding is ahead by 5 to 8 points where its training covered the render
style, while under full domain shift the readout, once its four thresholds
are calibrated, stays ahead. Within
real scans the readout's advantage grows with plan size and reverses on small
plans; neither ink density (the fraction of dark pixels, defined in
\cref{sec:emitdetect}) nor plan size explains the cross-domain reversal.
Openings show the same pattern as walls on real scans. The network
had always trained an opening heatmap that the decoder never read; reading it
raises opening F1 from $0.25$ to $0.64$ with no retraining
(\cref{sec:emitdetect}). The practical rule, stated empirically: detect on large real scans and
decode on small ones; on clean renders, decode where the decoder has trained
on the style, otherwise detect with in-domain-calibrated thresholds.

The question of readout is usually overshadowed by the question of output
representation. Room-centric vectorizers
\citep{yue2023roomformer,liu2024polyroom,xu2024frinet,phung2026raster2seq}
predict each room as a separate polygon and derive walls afterwards, which
appears to produce double walls and walls on open-plan boundaries; wall-first
serializations \citep{floorplanvlm2026,cage2025} make those errors impossible
to express and are credited with the resulting topological validity. This was
also our own starting assumption. In a comparison with matched data, recipe,
and supervision (\cref{sec:refute}), the representation turns out to be
secondary. The room-centric system, after a deterministic reconciliation step,
produces no double walls, reaches wall precision around $0.9$ and
watertightness of $0.92$--$0.94$, and has an edit cost about 30\% below native wall-first emission.
The failure modes we had attributed to the representation (wall precision
$0.22$ with double walls throughout, in an early zero-shot audit) are accounted
for by three factors that the controlled comparison removes, namely the
ground-truth convention, the reconciliation step, and domain shift; we did not
run an arm that isolates training supervision alone. The representational
effect that remains is a coverage limit: walls that bound no room cannot be recovered from
room output (recall $0.30$ vs.\ $0.66$), which accounts for about 4 points of
aggregate recall and says nothing about topological quality.

The two families are still complementary, and where that
complementarity can be used turns out to matter. Feeding one system's output
to the other as a conditioning input does not help: in three forms (a
rasterized room-outline channel, a rasterized fused-wall-draft channel, and a
symbolic coordinate prefix), run under a stop rule fixed in advance, no arm
improved on its zero-content control, and two controls that supply
ground-truth content moved the score by at most 0.3 points
(\cref{sec:cond}). A draft of this kind is another extractor's reading of the
same image, and a model that already sees the image gains nothing from it.
Combining the two outputs after prediction, by deterministic fusion of
room-derived and wall-detected walls, raises wall F1 by $7$ points over the
better system alone (paired 95\% interval $[5.9,8.3]$; \cref{sec:fusion}).

\paragraph{Contributions.}
(1)~A comparison of emitting and detecting on a fixed trained network, showing
that the better readout depends on plan size within real scans and on the
domain across domains, with a practical rule (detect on large real scans and decode on small ones;
on clean renders, decode where the decoder has trained on the style,
otherwise detect with in-domain-calibrated thresholds) and the observation
that, with edge chaining off, the readout's thresholds transfer across
domains for networks used in or finetuned to their training domain, while
under domain shift the thresholds are what calibration must supply
(\cref{sec:emitdetect}). (2)~A negative result on input-side
conditioning, with a stop rule fixed in advance and an explanation of why the
prior cannot add information, together with the observation that the same
prior is useful as output-level fusion (\cref{sec:fusion,sec:cond}). (3)~A
controlled study with matched data and recipe, showing that the
representation-first account of floorplan topological quality does not hold in
our setting: on wall precision, double walls, and watertightness, room emission
followed by reconciliation matches or exceeds native wall-first sequence
emission, and the failure modes previously attributed to the room-centric
representation are accounted for by the evaluation convention, the
reconciliation step, and domain shift (\cref{sec:refute}). (4)~An edit-cost
metric that scores drafts by the human work needed to correct them, with a
weight-sensitivity check, and corrected CubiCasa5K annotations (\emph{skv4})
with the correction tooling (\cref{sec:metrics,sec:bench,app:train}).
(5)~ResPlan-FP, a CC~BY~4.0 benchmark of 16{,}998 plans with frozen,
fingerprinted splits, a published render and evaluation protocol, and baseline
tracks for zero-shot, finetuned, and LLM systems (\cref{sec:resplan}).

% =====================================================================
\section{Related Work}
% =====================================================================
\paragraph{Floorplan vectorization.}
Early systems combined binarization, line extraction, and hand-written rules
\citep{mace2010rooms,ahmed2011improved,delasheras2014statistical}.
Raster-to-Vector \citep{liu2017raster} formulated the task as neural junction
detection followed by integer programming. Multi-task segmentation networks
\citep{kalervo2019cubicasa,zeng2019deepfloorplan} improved pixel labeling but
still rely on raster-to-vector post-processing, which has been criticized as
unreliable on non-Manhattan geometry \citep{floorplanvlm2026}. A second family
reconstructs floorplans from 3D scans or density maps rather than drawings:
Floor-SP \citep{chen2019floorsp}, FloorNet \citep{liu2018floornet}, MonteFloor
\citep{stekovic2021montefloor}, HEAT \citep{chen2022heat}, PolyDiffuse
\citep{chen2023polydiffuse}, and the edge-centric CAGE \citep{cage2025}. These
share our concern with watertightness, but their input is metric scan data,
noisy and incomplete as it may be, rather than a stylized drawing.

\paragraph{Two representational choices and the claims made for them.}
Learned vectorizers differ in what they emit. Room-centric methods predict each
room separately. RoomFormer's two-level queries \citep{yue2023roomformer},
PolyRoom's vertex queries \citep{liu2024polyroom}, and FRI-Net's per-room
implicit functions \citep{xu2024frinet} were developed for scan and density-map
input and were adapted to raster drawings as baselines by Raster2Seq.
Raster2Seq itself \citep{phung2026raster2seq} emits labeled polygon sequences
(rooms plus door and window polygons, no walls) directly from the raster image,
reports the strongest published room metrics, and generalizes to the in-the-wild WAFFLE
corpus \citep{ganon2025waffle}. Wall- and edge-centric methods emit wall or edge
structure directly. Raster-to-Graph \citep{hu2024rastertograph} predicts a
floorplan graph one vertex at a time. FloorplanVLM \citep{floorplanvlm2026}
fine-tunes a vision-language model to emit a dependency-ordered JSON (wall
skeleton first, then rooms as cycles of wall IDs) and attributes topological
validity to that serialization. CAGE \citep{cage2025} emits per-room edge
sequences, which is edge-centric but does not share walls between rooms, and
attributes validity to the edge formulation. These works support a
representation-first account of topological quality. Our controlled study
(\cref{sec:refute}) tests that account directly and does not confirm it: with
data and recipe held equal, the room-centric system matches or exceeds the
wall-first one on wall-structure metrics. We use both families as
experimental instruments rather than proposing either, and we locate our
contribution in the effects of evaluation convention and post-processing,
output-level fusion, and readout form, none of which is a change of
representation.

\paragraph{Conditioning one predictor on another's output.}
Feeding a structured predictor a prior from another source is a recurring
idea. PolyGen \citep{nash2020polygen} references vertex indices when emitting
faces, SceneScript \citep{avetisyan2024scenescript} references parent-wall IDs
when placing openings, and PolyDiffuse \citep{chen2023polydiffuse} denoises
room polygons produced by an existing method, conditioned on the scan.
\Cref{sec:cond} tests this kind of transfer between the two families in three input
forms (a raster room-outline channel, a raster fused-draft channel, and a
symbolic coordinate prefix) and finds none of them useful. The result was
obtained under a stop rule fixed in advance and includes two controls that
supply ground-truth content. The comparison with PolyDiffuse is informative,
although the interpretation is ours and not a claim PolyDiffuse makes: its
density-map inputs leave the geometry under-determined, so a conditioned
proposal has non-redundant work to do, whereas a clean floorplan drawing already
determines its walls, and a second description of the same evidence adds
nothing.

\paragraph{Structured sequences for visual geometry.}
Treating perception as sequence generation began with image captioning
\citep{vinyals2015show,xu2015showattend} and was generalized by Pix2seq
\citep{chen2021pix2seq,chen2022unified}. PolyFormer \citep{liu2023polyformer}
emits segmentation polygons as coordinate sequences, and PolygonRNN++
\citep{acuna2018polygonrnnpp}, which like our work is motivated by annotation
cost, generates polygons for interactive labeling. General-purpose VLMs that
emit free-text coordinates have been argued to fail on this task, because
token-by-token coordinate generation is probabilistic while the geometry is
exact, and because long structured outputs degrade \citep{floorplanvlm2026}. We
would add that vision encoders downsample thin wall lines away. Our zero-shot
VLM row (\cref{sec:resplan}) confirms the failure empirically: the topology is
roughly right and the coordinates are coarse. LLM-guided parsers
keep dedicated detectors in the loop \citep{ayanzadeh2026llmparse}. This
detector-versus-decoder split is the emit-versus-detect question we quantify
in \cref{sec:emitdetect}.

\paragraph{Floorplan datasets and synthetic data.}
Public corpora with vector labels are few. CubiCasa5K
\citep{kalervo2019cubicasa} has 5k Finnish plans under CC~BY-NC-SA~4.0;
Structured3D \citep{zheng2020structured3d} has 3.5k CAD scenes under a
non-commercial license; RPLAN \citep{wu2019rplan} has 80k generated layouts;
ResPlan \citep{resplan2025} has 17k vector graphs (data CC~BY~4.0, code MIT);
and WAFFLE \citep{ganon2025waffle} is large but has little geometric
annotation. None covers stylized marketing plans or labels open boundaries, and
the raster images used by Raster-to-Vector remain restricted (LIFULL/IDR,
institutional access only). This is why we build a redistributable benchmark on
ResPlan (\cref{sec:resplan}). Work on learned floorplan generation
\citep{li2019grains,hu2020graph2plan,sun2022wallplan,shabani2023housediffusion,paschalidou2021atiss}
aims at plausible new layouts rather than supervision pairs. We instead use
procedural generation in the spirit of domain randomization
\citep{tobin2017domainrandomization}: each generated vector plan is rendered
through randomized style and degradation layers, with the label transform known
exactly. Structured3D supplies real layouts for pretraining. We do
not evaluate on the Structured3D official test split, because our pretraining
predates the split freeze and overlaps it.

\paragraph{Evaluating vectorization.}
The standard protocol reports precision, recall, and F1 of rooms, corners, and
angles \citep{liu2017raster,chen2022heat,yue2023roomformer,phung2026raster2seq},
sometimes gated by IoU; sequence methods add a validity rate
\citep{floorplanvlm2026}. These measures are insensitive to the failure modes
that dominate correction effort, and they saturate at tolerances that hide
localization differences. We report F1 over a tolerance sweep, watertightness
read together with double-wall counts, and the edit-cost metric of
\cref{sec:metrics}. Interactive-annotation work has measured human effort as
clicks saved \citep{acuna2018polygonrnnpp}; our metric extends this from
polygon clicks to a typed editor-operation cost over walls, rooms, and
openings.

% =====================================================================
\section{Task, Systems, and Metrics}\label{sec:metrics}
% =====================================================================
\paragraph{Task and the two systems under test.}
The input is a rasterized floorplan. The output is a watertight vector plan of
walls (centerline segments), openings (door and window intervals on a wall),
and labeled rooms (polygons). We compare two systems rather than two
representations in isolation, since neither system is a pure representation:
\begin{itemize}
\item \textbf{Room-centric (Raster2Seq + reconciliation).} An autoregressive
  model emits labeled room polygons. A deterministic reconciliation step
  (welding, shared-edge deduplication, T-splitting, re-projection) derives walls
  from the polygon edges.
\item \textbf{Wall-first.} A grammar-masked autoregressive decoder emits walls
  as primitives, rooms as cycles of wall IDs with explicit open edges, and
  openings as intervals along a wall. A dense junction and centerline head snaps
  endpoints to obtain watertightness. The grammar rules out duplicate wall IDs, but near-coincident wall segments can still occur and have to be removed
  by snapping or merging (\cref{tab:postproc,tab:emit}).
\end{itemize}

\paragraph{fpeval.}
We score wall, room, and opening F1 with Hungarian matching at IoU$\ge$0.5 over
a tolerance sweep ($0.015$--$0.08$; matching criteria in \cref{app:fpeval}),
together with a watertightness rate
$r_{\mathrm{val}}$ (validity of the region tiling, allowing open edges). We
always read $r_{\mathrm{val}}$ together with the double-wall count, because a
plan made of duplicated walls closes trivially and $r_{\mathrm{val}}$ alone
would not reveal it. The room metric follows the official IoU@0.5 matching. To
check agreement with the official evaluator, we re-scored the official
Raster2Seq checkpoint on our local rebuild of its preprocessing and data split
and obtained a room F1 of $0.8854$ against the published $0.887$.

\paragraph{Edit cost.}
Our main metric scores a draft by the human effort needed to correct it into
the ground truth. Let $P$ be a predicted plan and $G$ the ground truth. We
first match predicted walls to ground-truth walls (Hungarian matching within a
move radius) and then read off an edit script $s(P,G)=(o_1,\dots,o_k)$ from the
matching: each matched pair becomes a \textsc{move} (plus a \textsc{type} change
if the wall types differ), each unmatched ground-truth wall a \textsc{create},
and each unmatched predicted wall a \textsc{delete}, or a \textsc{convert} if it
lies on a ground-truth open boundary. Openings are handled the same way with
their own three operation types. The cost is the weighted sum of this script,
\begin{equation}
  E_w(P,G)\;=\;\sum_{o_i\in s(P,G)} w_{\tau(o_i)}\,\ell(o_i),
  \label{eq:edit}
\end{equation}
where $\tau(o_i)\in\mathcal{T}=\{\textsc{move},\textsc{type},\textsc{create},
\textsc{delete},\textsc{convert}\}$ and $\ell(\cdot)$ is the magnitude of the
operation ($\ell=$ endpoint displacement in units of the tolerance for
\textsc{move}; $\ell=1$ otherwise). The script is determined by the matching
and is not a minimum over all possible scripts. The full procedure, the move
radius, and the opening operations are given in \cref{app:edit}. Our claim
concerns the partial order on $w$,
\begin{equation}
  w_{\textsc{move}}\;\ll\;w_{\textsc{type}}\;<\;w_{\textsc{create}}=w_{\textsc{delete}}
  \;\ll\;w_{\textsc{convert}},
  \label{eq:order}
\end{equation}
which follows the number of editor interactions each operation requires. A move
is one drag of an existing element; a type change is one property edit; create
and delete require locating and drawing geometry; and an open-to-phantom
\textsc{convert} also requires re-checking the topology of every adjacent room.
The magnitudes we use ($1,2,5,5,10$) are set by hand. \Cref{sec:bench} reports
that the ranking of systems does not change anywhere in the weight region
satisfying \cref{eq:order}, so the results depend on the ordering rather than
on the magnitudes. Calibration against logs from a deployed editor is left for
future work.

\paragraph{Leakage control.}
Every cross-generation evaluation asserts that the train and evaluation sets
share no plan id. All splits are frozen with id-hash fingerprints.

% =====================================================================
\section{The Wall-First Network and Its Two Readouts}\label{sec:method}
\begin{figure}[htbp]
\centering
\includegraphics[width=0.95\linewidth]{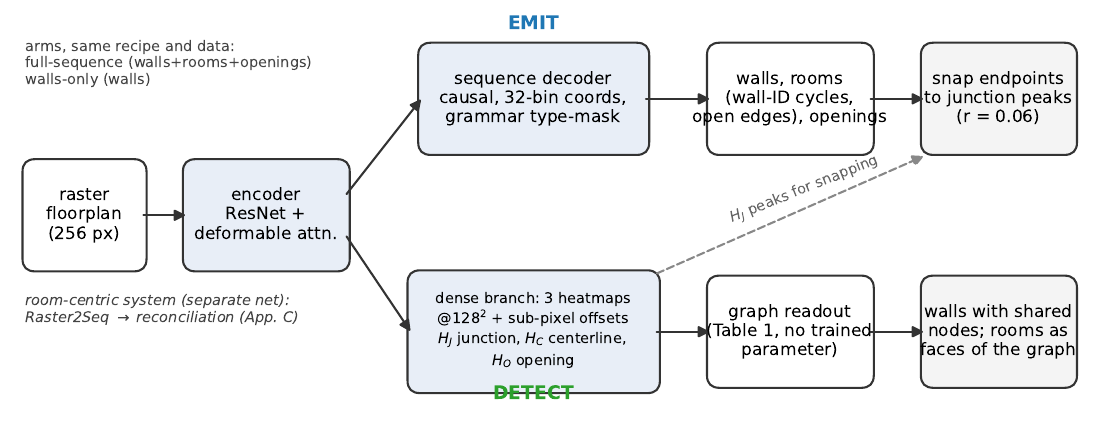}
\caption{The wall-first network and its two readouts. One encoder feeds a
grammar-masked sequence decoder (emit) and a dense branch of three heatmaps;
the graph readout (detect) assembles walls from the junction and centerline
heatmaps without a trained parameter. The room-centric system is a separate
network (Raster2Seq) followed by a reconciliation step.}
\label{fig:wfarch}
\end{figure}

% =====================================================================
The wall-first network is the instrument of \cref{sec:emitdetect,sec:cond},
one of the two systems of \cref{sec:refute}, and the donor of
\cref{sec:fusion}; \cref{fig:wfarch} shows it. Hyperparameters are in
\cref{app:train}.

\paragraph{Backbone and sequence decoder.} The network reuses the Raster2Seq
backbone (ResNet features and a deformable-attention encoder) on a $256$\,px
input, followed by a causal autoregressive decoder with anchor-based
coordinate embedding and $32$-bin coordinate quantization per axis. The output
grammar has three primitive types: a wall is a start point, an end point, a
thickness, and a type; a room is a cycle of wall IDs in which an edge may be
marked open (no wall); an opening is a door or window interval along a wall.
A type mask at every decoding step restricts the next token to what the
grammar allows at that position, so the decoder cannot emit a duplicate wall
ID or a room that references a wall it has not emitted. What the mask cannot
prevent is two near-coincident wall segments with different IDs; these are
removed by snapping or, in the room-centric system, by reconciliation.

\paragraph{Dense branch.} Alongside the sequence loss, three heatmaps are
trained at $128\times128$ with sub-pixel offsets: junctions, wall centerlines,
and openings. They are trained jointly with the sequence and share the
encoder.

\paragraph{Two readouts.} \emph{Emit}: the sequence decoder produces walls,
rooms, and openings; wall endpoints are then snapped to the nearest junction
peak within $0.06$ of the frame, which is what closes rooms. \emph{Detect}:
the graph readout of \cref{alg:readout} takes the peaks of the junction
heatmap as nodes and decides edges from centerline coverage; it has no trained
parameter, produces walls with shared node instances, and takes rooms as the
faces of the resulting planar graph. Its thresholds are calibrated on the
validation split of each domain (\cref{app:readout}). The opening heatmap can
be read the same way (\cref{sec:emitdetect}).

\paragraph{Arms.} The \emph{full-sequence} arm emits walls, rooms, and
openings in one sequence and is the wall-first row of \cref{tab:main}. The
\emph{walls-only} arm spends the whole sequence budget on walls; it is the
network whose two readouts \cref{tab:emit} compares, the base of the
conditioning arms in \cref{sec:cond}, and the donor in \cref{sec:fusion}.
Both are trained with the same recipe on the same data regimes
(\cref{sec:setup}).

\paragraph{Room-centric system.} Raster2Seq \citep{phung2026raster2seq},
continued from its released checkpoint on the same data regimes, emits
labeled room polygons and door and window polygons; the reconciliation step of
\cref{app:fpeval} welds vertices, de-duplicates shared edges, and T-splits to
derive walls.

% =====================================================================
\section{Experimental Setup}\label{sec:setup}
% =====================================================================
\textbf{Datasets.} CubiCasa5K real scans, for which we release corrected
\emph{skv4} annotations and the correction tooling (official split
$3328/318/296$); a procedurally generated synthetic set (\cref{app:synth}) whose
unit-level statistics are fitted to 18k real residential listings, with
several tiers including an open-plan tier that CubiCasa does not provide; and ResPlan-FP, our
clean vector-render benchmark (\cref{sec:resplan}). \textbf{Recipe.} All arms
share a frozen recipe (Structured3D real-layout initialization, 500 epochs,
dense heads, snap radius $0.06$) and are trained separately for each data
regime: \texttt{cc5k}, \texttt{cc5k+synth} ($1{:}1$, ``mix''), and
\texttt{synth-only}. \textbf{Leaderboard scope.} Official CubiCasa numbers are
cited rather than re-trained. Our system's rooms are the Raster2Seq stage-1
output, so an official-convention row would reproduce the published $0.887$.
Our contributions on walls, openings, edit cost, and watertightness are not
visible under the rooms-only official protocol; they appear in the controlled
tables and in fpeval.

% =====================================================================
\section{Evaluation}
% =====================================================================

% ---------------------------------------------------------------------
\subsection{Emit versus detect}\label{sec:emitdetect}
% ---------------------------------------------------------------------
We begin with the comparison that the title asks about: keeping the trained
network fixed and changing only the readout. The network is the walls-only
wall-first arm (\cref{sec:method}), not the full-sequence model of
\cref{tab:main}: the graph readout produces walls only, so the matched
comparison is against a decoder that spends its whole sequence budget on
walls. The full-sequence model's walls score $0.751$ at $t=0.05$ against
$0.790$ for the walls-only decoder on the same 296 plans (paired difference
$3.9$pp, 95\% interval $[2.9,5.0]$), which is the cost of emitting rooms and
openings in the same sequence. We compare autoregressive sequence decoding
with a graph readout of the same network's dense branch that requires no
training: peaks of the junction heatmap become nodes, and the centerline
heatmap's coverage of the chord between two nodes decides whether they are
joined by an edge (\cref{alg:readout}). All readout hyperparameters
($\theta_J,\theta_C,\kappa$, and whether to chain) are calibrated on the
validation set only, separately per domain (\cref{app:readout}).

\begin{table}[t]
\centering
\small
\caption{Graph readout from dense heatmaps to a wall graph. $H_J$ and $H_C$ are
the network's junction and centerline heatmaps; no parameter is trained.}
\label{alg:readout}
\begin{tabular}{p{0.94\linewidth}}
\toprule
\textbf{Input:} $H_J,H_C\in[0,1]^{S\times S}$, sub-pixel offsets $O$; thresholds
$\theta_J,\theta_C$, coverage $\kappa$, chain angle $\alpha$.\\
\textbf{1. Nodes.} $V\leftarrow$ local maxima of $H_J$ with $H_J\ge\theta_J$
(non-maximum suppression), refined by $O$.\\
\textbf{2. Candidate edges.} For every pair $(u,v)\in V^2$: sample the chord
$\overline{uv}$ at $n$ interior points (trimming the ends); accept if the
fraction of samples with $H_C\ge\theta_C$ is at least $\kappa$, $\|u-v\|\ge
\ell_{\min}$, and no third node lies within $\epsilon$ of the chord
(composite chords are rejected; their sub-edges carry them).\\
\textbf{3. Chain.} At each node, continue an edge into its single best
anti-parallel partner within angle $\alpha$; walk chains in both directions to
obtain maximal near-straight walls. Disabled in every frozen configuration: the
ground-truth walls of all three domains are split at junctions, and chaining
costs $22$pp on CubiCasa5K validation (\cref{app:readout}).\\
\textbf{Output:} walls as chained segments with shared node instances, hence
closed by construction; rooms as faces of the planar graph.\\
\bottomrule
\end{tabular}
\end{table}

\begin{figure}[p]
\centering
\includegraphics[width=\linewidth]{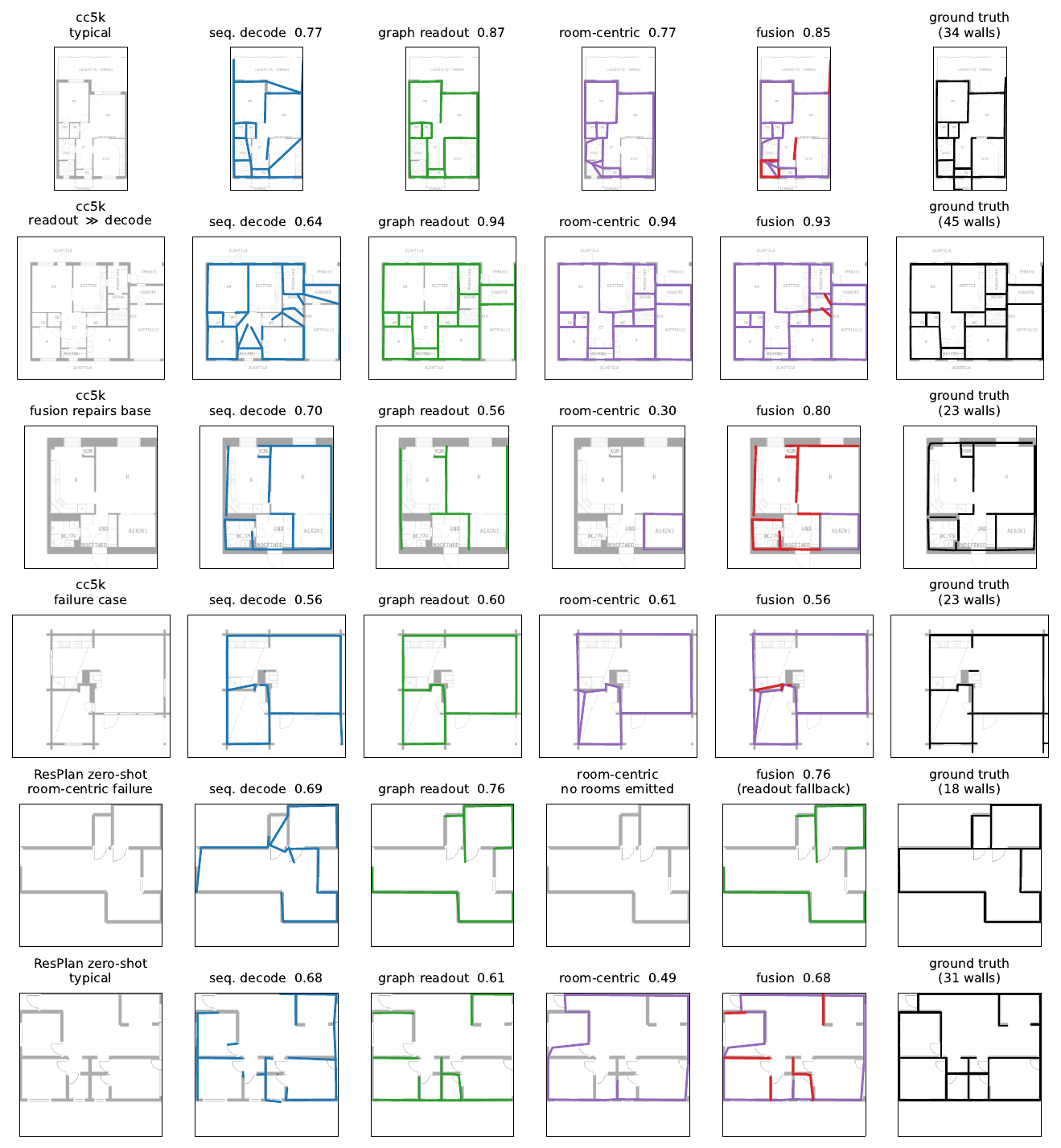}
\caption{Qualitative results. Columns: input; sequence decode (walls-only
arm); graph readout of the same network; room-centric system after
reconciliation; fusion (base in purple, ink-gated donor walls in red);
ground truth. Rows 1--4 are CubiCasa5K official-test plans (a typical plan, a
plan where the readout is far ahead of decoding, a plan where fusion repairs
a weak room-centric output, and a failure case on which every system is
poor); rows 5--6 are ResPlan-FP zero-shot plans, one on which the room-centric
arm emits no rooms and the fused system falls back to the readout walls. The
ResPlan readout panels use the frozen CubiCasa5K thresholds (the zero-shot protocol
of \cref{tab:resplan}), not the in-domain-calibrated variant. Wall F1 at
$t=0.05$ in each title.}
\label{fig:qual}
\end{figure}

\paragraph{On real scans, detection is better on every wall measure
(\cref{tab:emit}).} The graph readout is ahead of sequence decoding by $2.7$pp (percentage points of F1)
at tolerance $0.05$ (paired bootstrap 95\% interval $[1.4,4.0]$, $n=296$) and
by $5.1$pp at $0.015$ ($[3.7,6.4]$), with precision $8.3$pp
higher.\footnote{Paired means are computed from unrounded per-plan scores and
can differ from differences of rounded table entries by $0.1$pp.} Throughout
the paper, intervals are percentile bootstrap intervals over plans
($10{,}000$ resamples), and paired intervals resample the per-plan difference.
On this test set the half-width of a single system's interval is about
$0.012$--$0.018$ and that of a paired difference about $0.010$--$0.015$, so
differences below one point are not distinguishable from zero.
The gap is largest at the strict tolerance, which is what one expects if
autoregressive coordinate emission is the source of the loss. It is also the only
case in our experiments where a single system with native wall output beats
the room-centric system on walls at both tolerances ($+3.8$pp $[2.3,5.4]$ at
$0.05$ on the common 287-plan roster; the room-centric system is compared in
full in \cref{sec:refute}). Shared node instances give
$r_{\mathrm{val}}=0.956$ at $0.503$ double walls per plan (with about $25$ predicted walls per plan against $29$ in the ground
truth; to be read jointly,
per \cref{sec:metrics}), against $0.196$ for
snapped sequence emission. Closure by construction turns a continuous
coordinate problem into a discrete edge-recall problem that can be checked.
Because the readout replaces hundreds of sequential decoding steps with one
dense forward pass, it is also much cheaper at inference than the decoder, and as a
donor in the fused system of \cref{sec:fusion} it is interchangeable with the
sequence decoder. \Cref{fig:qual} shows both readouts, the room-centric
system, and the fused output on representative plans.

\begin{table}[t]
\centering
\caption{Same trained network (walls-only arm) with the wall readout swapped,
CubiCasa5K official test ($n{=}296$). Detection is ahead of autoregressive decoding
on every wall measure. Rooms are the faces of the wall graph; the readout emits
no openings. The room-centric system is shown for reference ($n{=}287$).
Per-system 95\% bootstrap half-widths on wall F1 are $0.012$--$0.018$.}
\label{tab:emit}
\small
\begin{tabular}{lccccccc}
\toprule
readout & F1@.05 & F1@.015 & P & R & $r_{\mathrm{val}}$ & rooms & open \\
\midrule
graph readout (detect)   & \tbf{0.818} & \tbf{0.787} & 0.885 & 0.775 & \tbf{0.956} & 0.591 & --- \\
sequence decode (emit)   & 0.790 & 0.736 & 0.802 & \tbf{0.787} & 0.196 & 0.512 & 0.391 \\
r2s\_mix (rooms+reconcil.)   & 0.781 & 0.750 & \tbf{0.906} & 0.709 & 0.916 & \tbf{0.784} & \tbf{0.848} \\
\bottomrule
\end{tabular}
\end{table}

\paragraph{Within real scans the winner depends on plan size
(\cref{tab:crossover}).} We had proposed that the density of ink explains
where the readout wins: dense heatmaps would have enough to respond to on a
scanned drawing and little on a thin line render. ``Ink density'' is not a
standard quantity, so we fix it operationally: the fraction of pixels of the
$256$\,px input whose gray value is below $100$ of $255$ (dark-pixel
fraction). We also split it into \emph{wall ink}, the same fraction restricted
to a band of the annotated thickness around each ground-truth wall, and
\emph{clutter}, the dark pixels outside those bands (furniture, text,
hatching). Within CubiCasa5K the readout's advantage does correlate with
dark-pixel fraction (Spearman $\rho=-0.37$, $p<10^{-10}$, $n=296$; wall ink
$-0.21$, clutter $-0.28$), but dark-pixel fraction is itself confounded with
plan size ($\rho=-0.47$ against the number of ground-truth walls; large plans
do have thinner walls relative to the frame: $\rho=-0.71$ between a plan's
mean annotated wall thickness and its wall count, with mean thickness $16.2$
frame units in the small tercile against $8.5$ in the large). Controlling for size, the ink
effect shrinks to a partial $\rho=-0.22$ while the size effect remains
$+0.34$ ($\rho=+0.46$ uncontrolled). Split by ground-truth wall count, the
readout \emph{loses} on small plans and wins by a wide margin on large ones:
$-3.7$pp on plans with 5--22 walls (paired 95\% interval $[-6.2,-1.2]$,
$n=97$), $+2.4$pp on 23--31 walls ($[0.6,4.3]$, $n=91$), and $+8.7$pp on
32--70 walls ($[7.1,10.4]$, $n=108$). The two readouts move in opposite
directions: readout F1 rises with plan size ($\rho=+0.22$) while decoder F1
falls ($\rho=-0.28$), which is what one expects if autoregressive decoding
degrades with sequence length and the dense readout does not. The practical
rule on real scans is therefore to detect on large plans and decode on small
ones, not to detect on scans as such.

\paragraph{Across domains, neither ink nor size explains the reversal.} On
clean synthetic renders (the open-plan tier) and clean vector renders
(ResPlan), sequence decoding is ahead by $5$ to $8$ points once finetuned or
calibrated in-domain (\cref{tab:crossover}). The ink premise was
wrong about the input: ResPlan-FP renders median-thickness wall \emph{bands},
not thin lines (\cref{app:resplan}), and the dark-pixel fraction is not
monotone in the winner (CubiCasa5K $0.078$, ResPlan $0.094$, open-plan $0.112$; wall
ink $0.56$, $0.73$, $0.52$). Plan size does not explain it either: the clean
domains have \emph{larger} plans (mean ground-truth walls $28.9$, $31.0$,
$35.0$), on which the readout should be favored. Two further probes locate the reversal. Zero-shot, the order
depends only on the readout's thresholds: carried over frozen, the walls-only
network decodes better than it detects on ResPlan-FP ($0.640$ against
$0.621$; paired $-1.9$pp $[-2.7,-1.2]$, $n=1000$), but calibrating just the
four readout thresholds on ResPlan validation, with no training, flips the
order ($0.677$ against $0.640$; $+3.7$pp $[+3.0,+4.4]$). Under full domain
shift the calibrated readout stays ahead, as on real scans. The
decoder's advantage appears where its training covered the render style: on
the open-plan tier, whose marketing render style is part of the training mix,
it leads the in-domain-calibrated readout by $8.2$pp ($[4.7,12.3]$, $n=30$),
and after finetuning on ResPlan it leads by $5.3$pp ($[5.0,5.5]$), the finetune having moved the decoder by $+32.8$pp and the readout by
$+23.8$.\footnote{Each increment mixes two variables: the decoder's increment spans checkpoints (final to best-validation) and the readout's spans threshold
baselines (zero-shot-calibrated to finetuned-calibrated). Holding the frozen
CubiCasa5K thresholds at both ends gives the readout $+29.3$pp ($0.621\to0.914$);
the decoder still gains more, though the margin narrows from $9.0$ to
$3.5$pp.}
The clean domains reward not emission as such but the decoder's
learned prior over the render style, which only training supplies; why that
prior beats dense detection on clean geometry where it is available remains
untested. Threshold sensitivity splits the same way: in the training domain
or after finetuning, the frozen CubiCasa5K thresholds transfer at a cost of at most
$0.6$pp ($0.914$ against $0.915$ on finetuned ResPlan, $0.819$ against
$0.825$ on the open-plan tier), while under domain shift the thresholds are
the fragile part ($0.621$ frozen against $0.677$ calibrated;
\cref{app:readout}). We report the rule as empirical: detect on large real
scans; decode on clean renders when the decoder has trained on the style,
otherwise detect with in-domain-calibrated thresholds; keep edge chaining off
everywhere (\cref{app:readout}).

\begin{table}[t]
\centering
\caption{Emit versus detect across domains (wall F1@.05) with the mean
dark-pixel fraction of the $256$\,px input and the paired bootstrap 95\%
interval of readout minus decode. The walls-only arm is used everywhere except
the finetuned ResPlan row. Readout thresholds: CubiCasa5K frozen; open-plan calibrated on an in-domain set
($0.819$ with the frozen CubiCasa5K thresholds); finetuned ResPlan calibrated on
ResPlan validation ($0.914$ with the frozen CubiCasa5K thresholds); the two
zero-shot rows show the frozen thresholds carried over and the effect of
calibrating only the thresholds on ResPlan validation (no training). The
winner is not monotone in ink density or in plan size.}
\label{tab:crossover}
\small
\setlength{\tabcolsep}{4.5pt}
\begin{tabular}{lcccccc}
\toprule
domain & $n$ & dark-pixel frac. & GT walls & seq.\ decode & graph readout & readout $-$ decode \\
\midrule
CubiCasa5K (scans)              & 296  & 0.078 & 28.9 & 0.790 & \tbf{0.818} & $+2.7$ $[1.4,4.0]$ \\
open-plan (clean synth)        & 30   & 0.112 & 35.0 & \tbf{0.908} & 0.825 & $-8.2$ $[-12.3,-4.7]$ \\
ResPlan, zero-shot (frozen thr.)      & 1000 & 0.094 & 31.0 & \tbf{0.640} & 0.621 & $-1.9$ $[-2.7,-1.2]$ \\
ResPlan, zero-shot (calibrated thr.)  & 1000 & 0.094 & 31.0 & 0.640 & \tbf{0.677} & $+3.7$ $[+3.0,+4.4]$ \\
ResPlan, finetuned             & 1000 & 0.094 & 31.0 & \tbf{0.968} & 0.915 & $-5.3$ $[-5.5,-5.0]$ \\
\bottomrule
\end{tabular}
\end{table}

\paragraph{Openings show the same pattern, without retraining.} The wall-first
network had always trained an opening heatmap that the sequence decoder never
used. Reading openings from that heatmap raises opening F1 from $0.246$
(sequence-emitted) to $0.644$ with a simple door/window family heuristic, and
to $0.799$ when the family is ignored, a $2.6\times$ improvement from the
readout alone (CubiCasa5K official test, $n=287$, scored with an internal matcher
stricter than fpeval, so these three numbers are comparable with each other
and not with \cref{tab:main}). A small standalone door/window
head ($200$K parameters) reaches $0.988$ family accuracy and closes the
classification gap; routed through it, the fused system's openings reach
$0.849$ against $0.848$ for the room-centric arm alone, that is, break-even
(\cref{sec:fusion}). Openings are a local detection problem
rather than a sequence-emission problem, and on this data the room-centric
system already solves it.

% ---------------------------------------------------------------------
\subsection{Representation is not the main factor}\label{sec:refute}
% ---------------------------------------------------------------------
\Cref{tab:main} gives the matched-data, matched-recipe comparison on the
CubiCasa5K official test set. The claim that wall-first emission produces
better structure than room-centric emission is not supported on the
wall-structure measures. The room-centric system, after reconciliation,
produces no double walls, reaches wall precision around $0.9$ and
watertightness of $0.92$--$0.94$, and has an edit cost about 30\% lower than native wall-first
sequence emission; on wall F1 it leads by $3.1$pp (paired 95\% interval
$[1.3,4.9]$, $n=287$).

Two qualifications bound this result. First, the two systems are not equally
mature. The room-centric arm continues a published system from its released
checkpoint, whereas the wall-first arm is our own, and its room and opening
outputs ($0.37$--$0.40$ and $0.34$--$0.37$) are clearly under-tuned. The
comparison therefore shows that room emission with reconciliation is not
behind this wall-first implementation on wall structure; it does not show that
a better wall-first model could not lead, and we make no claim on rooms or
openings, where part of the room-centric lead is maturity. Second, the earlier
evidence for the representation-first account came from a zero-shot audit of
the released Raster2Seq checkpoint on a different domain (stylized marketing
plans), where the derived walls had precision $0.22$ with double walls
throughout. Three factors separate that audit from \cref{tab:main}, and none
of them is the representation: (i)~the ground-truth convention, since
re-scoring a fixed checkpoint against the corrected annotations alone moves
wall F1 from $0.74$ to $0.825$ (\cref{app:train}); (ii)~the reconciliation
step, since without it the same model has $7.9$ double walls per plan and
$14$pp lower precision (\cref{tab:postproc}); and (iii)~domain shift. An arm
that varies training supervision alone while holding the rest fixed was not
run, so we attribute the earlier failure to these three factors jointly and
not to training supervision specifically.

\begin{table}[t]
\centering
\caption{Matched-data comparison, CubiCasa5K official test, fpeval @0.05.
Room-centric rows are scored on the $290$ (r2s\_cc5k) and $287$ (r2s\_mix)
plans on which the model produces a decodable output, wall-first rows on all
$296$; on the common $287$-plan roster wf\_mix has wall F1 $0.751$ and edit
cost $110.2$ (\cref{tab:editops}). ``rooms'' for wall-first is derived from
faces; room-centric rooms are the Raster2Seq stage-1 output scored under fpeval
against skv4, not under the official protocol (which gives $0.887$,
\cref{sec:setup}). Per-system 95\% bootstrap half-widths on wall F1 are
$0.013$--$0.017$. Wall-first has higher wall recall; the room-centric system is
better on wall F1, rooms, openings, watertightness, and edit cost.}
\label{tab:main}
\small
\begin{tabular}{lcccccc}
\toprule
system & wall F1 & wall P & wall R & rooms & open & edit \\
\midrule
r2s\_cc5k (rooms+reconcil.)  & 0.769 & \tbf{0.894} & 0.703 & \tbf{0.777} & 0.840 & \tbf{80.7} \\
r2s\_mix  (rooms+reconcil.)  & \tbf{0.781} & \tbf{0.906} & 0.709 & \tbf{0.784} & \tbf{0.848} & \tbf{79.3} \\
wf\_cc5k  (full seq)     & 0.747 & 0.762 & 0.742 & 0.374 & 0.341 & 117.1 \\
wf\_mix   (full seq)     & 0.751 & 0.764 & \tbf{0.746} & 0.395 & 0.367 & 113.9 \\
\bottomrule
\end{tabular}
\end{table}

\begin{figure}[htbp]
\centering
\includegraphics[width=0.72\linewidth]{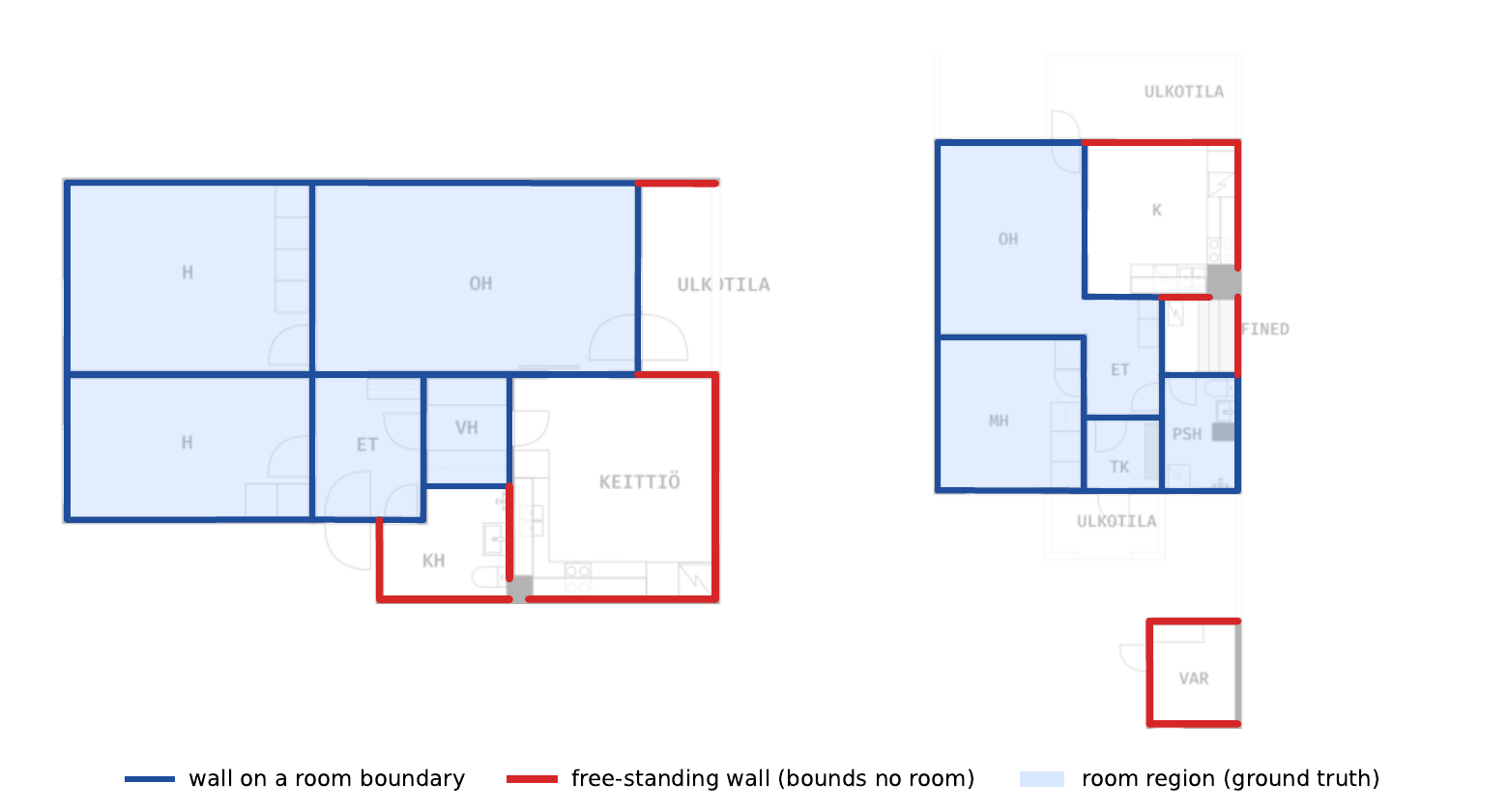}
\caption{Wall classes. Room-bounding walls can be recovered from room
polygons; non-room-bounding walls (dangling, stub, exterior) cannot.}
\label{fig:wallclasses}
\end{figure}

\paragraph{The remaining representational effect is a coverage limit.}
We split the ground-truth walls by whether they lie on any ground-truth room
cycle (\cref{tab:envelope}, \cref{fig:wallclasses}). On walls that the room
representation can express, the room-centric system is in fact more accurate
($+3.8$pp on boundary walls). The whole of the aggregate recall gap comes from
walls that bound no room (dangling, stub, and exterior walls), which room
polygons cannot carry. Writing aggregate recall as a mixture over the two wall
classes,
\begin{equation}
  R \;=\; \pi\,R_{\mathrm{b}} + (1-\pi)\,R_{\mathrm{nb}},
  \qquad \pi=\tfrac{3337}{3337+835}=0.80,
  \label{eq:envelope}
\end{equation}
gives $R_{\mathrm{r2s}}=0.80\cdot0.865+0.20\cdot0.302=0.752$ and
$R_{\mathrm{wf}}=0.80\cdot0.827+0.20\cdot0.663=0.794$, a gap of $4.2$ points
that comes entirely from the $(1-\pi)R_{\mathrm{nb}}$ term, even though the
room-centric system leads on $R_{\mathrm{b}}$. This is a limit on what the
representation can express, not a difference in topological quality.

\begin{table}[t]
\centering
\caption{Wall recall by room-boundary membership (150-plan subset of the CubiCasa5K
official test, tol 0.05). The representational effect is a coverage limit
rather than a topology-quality effect.}
\label{tab:envelope}
\small
\begin{tabular}{lcc}
\toprule
arm & room-boundary walls ($n{=}3337$) & non-boundary walls ($n{=}835$) \\
\midrule
r2s (rooms+reconcil.)   & \tbf{0.865} & 0.302 \\
wall-first native   & 0.827 & \tbf{0.663} \\
\bottomrule
\end{tabular}
\end{table}

\paragraph{Without post-processing the ordering reverses.} A symmetric
ablation (\cref{tab:postproc}) compares naive r2s, in which every room-polygon
edge becomes a wall, with wall-first emission with endpoint snapping turned
off. Each system is better on its own native output type: wall-first beats
naive r2s on walls, and r2s is better on rooms. The overall advantage of r2s
comes from the direction of derivation rather than from emission quality.
Deriving walls from rooms is a local operation and close to lossless (welding
and deduplication reduce $7.9$ double walls per plan to $0$ and raise precision
by $14$pp; note that the naive variant has the \emph{higher}
$r_{\mathrm{val}}$, $0.983$ against $0.938$, which is exactly the double-wall
inflation of watertightness that \cref{sec:metrics} warns about), whereas deriving rooms from walls is global, since every cycle has
to close, and loses information. The snap radius needed to close rooms lowers
wall-first wall F1 by about $3$pp.

\begin{table}[t]
\centering
\caption{Post-processing ablation, CubiCasa5K test, fpeval @0.05. Without the
reconciliation step or snapping, each system is better on its own native
element; the aggregate ordering is set by the asymmetry of derivation
direction.}
\label{tab:postproc}
\small
\begin{tabular}{lccccc}
\toprule
system & wall F1 & wall P & rooms & doubles/plan & $r_{\mathrm{val}}$ \\
\midrule
r2s\_cc5k + reconcil.       & 0.769 & \tbf{0.894} & \tbf{0.777} & \tbf{0.00} & 0.938 \\
r2s\_cc5k NAIVE         & 0.708 & 0.758 & 0.778 & 7.90 & 0.983 \\
wf\_cc5k snap .06       & 0.747 & 0.762 & 0.374 & $\sim$0 & 0.02 \\
wf\_cc5k SNAP0          & \tbf{0.778} & 0.794 & 0.087 & 0.70 & 0.01 \\
\bottomrule
\end{tabular}
\end{table}

% ---------------------------------------------------------------------
\subsection{Output-level fusion}\label{sec:fusion}
% ---------------------------------------------------------------------
The two systems are complementary, but the complementarity is in their
errors. A deterministic fusion keeps the room-centric reconciled walls as they
are and adds wall-first walls that do not overlap them and that pass an
\emph{ink gate}: the fraction of points sampled along the segment that have a
dark pixel within a few pixels perpendicular to it must be at least $0.5$
(\cref{app:fpeval}). This reaches $0.853$ wall F1 at
tolerance $0.05$, $0.811$ at $0.015$, and an edit cost of $73.1$
(\cref{tab:fusion}); the gain over the
room-centric system alone is $7.1$pp at $0.05$ (paired 95\% interval
$[5.9,8.3]$, $n=287$) and $6.2$pp at $0.015$ ($[5.1,7.2]$), and the edit-cost
reduction is $6.3$ ($[4.9,7.6]$). The ladder in \cref{tab:fusion} states the
donor in each row: the first two fusion rows use the full-sequence wf\_mix
model as donor, and the final system uses the walls-only arm of
\cref{sec:emitdetect} read by its sequence decoder, which is purer
(\cref{tab:purity}). Substituting the graph readout of the same network as
donor gives $0.850$ / $0.813$ (paired differences $-0.3$ and $+0.2$pp, both
intervals containing zero), so the two readouts are interchangeable as
donors; ungated, the readout donor reaches $0.856$ / $0.818$
(\cref{tab:purity}). The three donor variants lie within $0.6$pp of one
another, and we keep the walls-only sequence donor as the designated final
system because its opening and edit-cost accounting is complete, not because
it is distinguishable on wall F1. Openings and rooms are taken
unchanged from the room-centric stage-1 output, and donor walls are added
without their openings (\emph{opening-strip}); including the wall-first arm's
openings would bring in its poor opening quality ($0.26$--$0.37$).

\begin{table}[t]
\centering
\caption{Fusion steps, CubiCasa5K official test ($n{=}287$ of 296), wall F1, with
the donor stated per row. Each step requires no training and is deterministic.
Opening-strip changes openings and edit cost only, not wall F1. The final
system is the best overall (paired 95\% interval of its gain over r2s\_mix:
$[5.9,8.3]$pp at $0.05$, $[5.1,7.2]$ at $0.015$).}
\label{tab:fusion}
\small
\begin{tabular}{llccc}
\toprule
system & donor & F1@.05 & F1@.015 & edit \\
\midrule
r2s\_mix alone                       & --- & 0.781 & 0.750 & 79.3 \\
wf\_mix alone (full sequence)        & --- & 0.751 & 0.694 & 110.2 \\
walls-only arm alone (seq.\ decode)  & --- & 0.790 & 0.736 & --- \\
fusion, ungated                      & wf\_mix & 0.832 & 0.787 & --- \\
fusion + ink gate ($\ge0.5$)         & wf\_mix & 0.846 & 0.806 & 80.1 \\
\tbf{fusion final} (ink gate, opening-strip) & walls-only & \tbf{0.853} & \tbf{0.811} & \tbf{73.1} \\
fusion (ink gate, opening-strip)     & graph readout & 0.850 & 0.813 & 72.2 \\
fusion (no gate, opening-strip)      & graph readout & 0.856 & 0.818 & --- \\
\midrule
same-representation ensemble control & wf\_mix + walls-only & 0.794 & 0.742 & --- \\
\bottomrule
\end{tabular}
\end{table}

\paragraph{How much of this is ensembling?} Two controls indicate that the
gain is only partly due to the difference in representation. An ensemble of two
wall-emitters (the full-sequence wf\_mix model and the walls-only arm) under
the same fusion operator reaches $0.794$, which is $0.4$pp above its better
member (walls-only, $0.790$), so cross-representation fusion is $5.9$pp above
this control. However, an ensemble of two r2s models trained on different data, with
the same representation, already recovers $5.8$pp of the $6.9$pp
cross-representation gain measured on that roster (\cref{app:rosters}; the
$287$-plan figure is $7.1$pp). Most of the improvement comes from
combining models with different errors, with a cross-representation component
of about $1$pp and an asymmetry of roles: the high-precision reconciled output
is the appropriate base and the high-recall detector the appropriate donor. We
describe the gain accordingly, as ensembling with a small cross-representation
component, rather than as representation complementarity.

\paragraph{The sign of the fusion gain has a closed form.}
Let the base system have $\mathrm{TP},\mathrm{FP},\mathrm{FN}$ counts and
$F_1=2\mathrm{TP}/(2\mathrm{TP}+\mathrm{FP}+\mathrm{FN})$. The donor contributes
$k$ non-overlapping walls of which a fraction $p$ (its purity) are correct;
those $pk$ walls turn misses into hits and the rest become false positives:
\begin{equation}
  F_1' \;=\; \frac{2(\mathrm{TP}+pk)}{2\mathrm{TP}+\mathrm{FP}+\mathrm{FN}+k}
  \quad\Longrightarrow\quad
  F_1' > F_1 \;\iff\; p \;>\; \frac{\mathrm{TP}}{2\mathrm{TP}+\mathrm{FP}+\mathrm{FN}}
  \;=\; \frac{F_1}{2}.
  \label{eq:fusion}
\end{equation}
A donor therefore helps exactly when its purity exceeds half the base $F_1$,
and the size of the gain grows with the number of correct additions $pk$,
which is bounded by the base's misses $\mathrm{FN}$, that is, by its recall
deficit. Both quantities vary with the domain, and \cref{tab:purity} reports
them where the added walls can be identified. On real scans the base $F_1$ is
$0.78$ (threshold $0.39$); the ink-gated donor adds $4.9$ walls per plan at
purity $0.69$ ($0.56$ before the ink gate, which is why the gate helps), far
above threshold, and the gain is $+7.1$; the full-sequence wf\_mix donor is
less pure ($0.51$ ungated, $0.67$ gated) and gains less. The graph readout as
donor is purer still ($0.74$) and needs no gate: ungated it gives $0.856$
against $0.850$ gated, because its chord-coverage test already verifies ink.
One caveat applies to the gate's measured benefit on CubiCasa5K: the skv4
ground truth was itself filtered with an ink-coverage gate
(\cref{app:train}), so a donor wall without ink under it is penalized by a
criterion correlated with the one that removed such walls from the
annotation. On the open-plan tier the base has
recall $0.92$ and $F_1$ $0.95$ (threshold $0.47$); the readout donor adds
$2.7$ walls per plan at purity $0.15$, and fusion loses $2.5$ points ($-1.4$
with the sequence donor). On ResPlan zero-shot the base is weak ($F_1$ $0.52$ on the plans where it
decodes, threshold $0.26$); the readout donor, carrying the frozen CubiCasa5K
thresholds, adds $9.1$ walls per plan at purity $0.82$, and fusion gains
$19.3$ points on those plans.
Openings fused from the wall-first arm have $p\approx0.3$ against a base $F_1$
of $0.85$ and give $-2.6$; routing them through the dedicated door/window head
raises $p$ to break-even. In-domain ResPlan fusion of the finetuned arms is net
negative and breaks watertightness; its purity was not measured. In every
measured case the sign of the gain matches the side of the threshold on which
the purity falls, and the practical consequence is that fusion should be enabled per domain and the donor gated on purity. \Cref{eq:fusion}
is a statement about pooled counts: applied to the pooled base counts of the
$287$-plan roster it reproduces the ungated rows of \cref{tab:purity} to
within $0.1$pp and overestimates the gated rows by at most $0.9$pp; the base
$F_1$ and threshold quoted in the text are per-plan means, which run below
the pooled micro-$F_1$ ($0.807$, threshold $0.40$).

\begin{table}[t]
\centering
\caption{Measured donor purity $p$ against the threshold $F_1/2$ of
\cref{eq:fusion}, wall F1@.05. $k$ is the mean number of ink-gated donor walls
added per plan; $p$ is the pooled fraction of those that match a ground-truth
wall the base had missed. CubiCasa5K rows are on the $287$-plan roster; ResPlan is
scored on the $733$ zero-shot plans where the room-centric base decodes.}
\label{tab:purity}
\small
\begin{tabular}{llcccccc}
\toprule
domain & donor & base $F_1$ & $F_1/2$ & $k$ & $p$ & fused $F_1$ & $\Delta$ \\
\midrule
CubiCasa5K & wf\_mix seq, no gate & 0.781 & 0.391 & 7.4 & 0.51 & 0.832 & $+5.1$ \\
CubiCasa5K & wf\_mix seq, ink gate & 0.781 & 0.391 & 4.8 & 0.67 & 0.846 & $+6.5$ \\
CubiCasa5K & walls-only seq, no gate & 0.781 & 0.391 & 7.1 & 0.56 & 0.844 & $+6.3$ \\
CubiCasa5K & walls-only seq, ink gate & 0.781 & 0.391 & 4.9 & 0.69 & 0.853 & $+7.1$ \\
CubiCasa5K & graph readout, no gate & 0.781 & 0.391 & 4.8 & 0.75 & 0.856 & $+7.4$ \\
CubiCasa5K & graph readout, ink gate & 0.781 & 0.391 & 4.5 & 0.74 & 0.850 & $+6.9$ \\
open-plan & graph readout, ink gate & 0.946 & 0.473 & 2.7 & 0.15 & 0.922 & $-2.5$ \\
ResPlan zero-shot & graph readout (frozen thr.), ink gate & 0.523 & 0.261 & 9.1 & 0.82 & 0.715 & $+19.3$ \\
\bottomrule
\end{tabular}
\end{table}

% ---------------------------------------------------------------------
\subsection{Input-side conditioning (\texorpdfstring{$0/3$}{0/3} under a pre-specified stop rule)}\label{sec:cond}
\begin{figure}[htbp]
\centering
\includegraphics[width=0.95\linewidth]{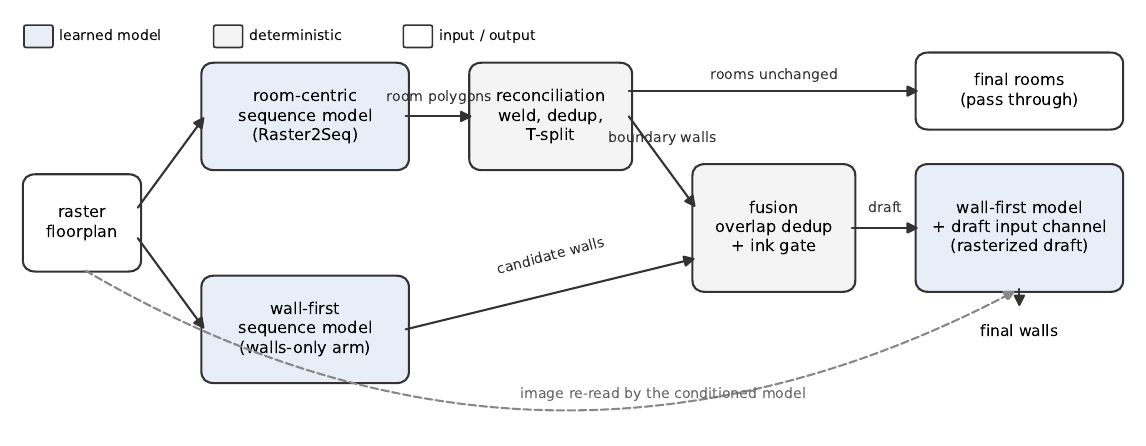}
\caption{The fused-draft-channel conditioning arm. The rasterized draft
enters the wall-first network as an extra input channel; the room-outline arm
replaces the draft raster with rasterized room outlines, and the
symbolic-prefix arm instead feeds the draft's coordinates to the decoder as a
prefix (not drawn). What any of them can carry is a second reading of the
same pixels, and none has an effect.}
\label{fig:arch}
\end{figure}

% ---------------------------------------------------------------------
Since the two outputs combine usefully after prediction, one might expect that
feeding one system's output to the other as an input would also help. We
fixed three conditioning forms and a stop rule in advance: three failures would
close the line of work (\cref{tab:cond}; architecture in \cref{fig:arch}). The
forms were a rasterized room-outline channel, a rasterized fused-wall-draft
channel, and a symbolic coordinate prefix. All three failed. Each scored at or
below its own zero-content baseline, and two controls that supplied
ground-truth content moved the score by at most $0.3$ points.

The explanation we propose is that such a draft is another extractor's reading
of the same image. Let $X$ be the image, $Y$ the target walls, and $D=g(X)$ the
draft produced by a deterministic extractor (greedy decoding of the other arm).
Then $H(D\mid X)=0$, hence
\begin{equation}
  I(Y;\,D\mid X)\;\le\;H(D\mid X)\;=\;0,
  \label{eq:cond}
\end{equation}
so a predictor that already sees $X$ cannot gain information from $D$. This is
a statement about information and not about optimization: a network of finite
capacity trained by gradient descent can benefit computationally from a
proposal it could in principle compute itself, which is how iterative
refinement and distillation work, so \cref{eq:cond} does not rule out a gain
from conditioning; it says that any gain must come from easier optimization
rather than from new evidence. Whether such a gain exists here is what the
arms test, and they find none. At best
$D$ acts as a regularizer; at worst, as here ($-2$ to $-3$ points), it is a
distraction whose error distribution at training time differs from that at
deployment. Ground-truth content is the complementary probe: $Y$ is not a
function of $X$, so $I(Y;Y\mid X)=H(Y\mid X)$, and the ground-truth arm has
headroom equal to the model's remaining uncertainty about the walls given the
pixels. That this headroom is at most $0.3$ points suggests that a clean
floorplan drawing leaves $H(Y\mid X)$ small: the ink already determines the
walls. This holds provided the conditioning pathway can read the content at
all. The training-time content of the channel bounds the alternative explanation
that the model simply learned to ignore a noisy channel: the room-outline arm
was trained on ground-truth rooms (with vertex jitter and room dropout)
throughout, and the draft-channel and prefix arms saw ground-truth walls with
synthetic corruption in $20\%$ of training samples ($60\%$ cross-arm drafts,
$20\%$ empty channel). A mid-training probe of the room-outline arm found no
reliance on the channel even so. No arm was trained on clean ground-truth
content alone, and that residual is not separated (\cref{sec:limits}). This is also how we read
the comparison with PolyDiffuse: density maps leave geometry under-determined
($H(Y\mid X)$ large, so conditioned proposals help), and clean floorplan ink
does not.

\begin{table}[t]
\centering
\caption{Input-side conditioning, CubiCasa5K official test, wall F1 (@.05 / @.015).
The three columns for each arm are deployment / zero-content /
ground-truth-content. No deployment arm beats the unconditioned wall-only model
(0.790 / 0.736); the zero- and GT-content control rows reproduce that baseline
within noise, and GT content adds at most $0.3$pt over zero-content. The
paired noise scale on this test set is about $\pm1.3$pt (\cref{sec:emitdetect}),
so the $0.3$pt control differences are indistinguishable from zero while the
$2$--$3$pt deploy-arm deficits are not.}
\label{tab:cond}
\small
\begin{tabular}{lccc}
\toprule
arm & deploy & zero-content & GT-content \\
\midrule
no-conditioning (walls-only) & \multicolumn{3}{c}{0.790 / 0.736} \\
room-outline channel         & 0.762 / 0.708 & 0.762 / 0.710 & 0.764 / 0.709 \\
fused-draft channel          & 0.768 / 0.720 & 0.792 / 0.742 & 0.795 / 0.745 \\
symbolic coordinate prefix   & 0.756 / 0.696 & 0.783 / 0.717 & 0.781 / 0.715 \\
\bottomrule
\end{tabular}
\end{table}

% ---------------------------------------------------------------------
\subsection{Edit cost}\label{sec:bench}
% ---------------------------------------------------------------------
On the edit-cost metric the fusion system is the best overall ($73.1$, against
$79.3$ for the room-centric system and $110.2$ for wall-first on the common
$287$-plan roster; bootstrap 95\% intervals $[68.8,77.5]$ and $[75.0,83.9]$
for the first two, and a paired fusion-minus-room-centric difference of $-6.3$
with interval $[-7.6,-4.9]$). The breakdown by
operation (\cref{tab:editops}, \cref{app:edit}) shows where the difference
lies. The wall-first system needs about three more wall deletions per plan
(extra and near-duplicate walls) and about six more opening creations and
deletions, while it needs six fewer type changes; fusion reduces wall creations
from $7.5$ to $4.0$ per plan at the price of $1.4$ more deletions. No
open-to-phantom conversion occurs on CubiCasa5K, because its ground-truth rooms
are all wall-bounded; that operation only matters on the open-plan tier. The
operation ordering is the claim; the magnitudes are set by hand. To check that the magnitudes do not drive the
result, we re-scored the three systems under $56$ weight vectors spanning
$w_{\textsc{move}}\in\{0.5,1,2\}$, $w_{\textsc{type}}\in\{1.5,2,3\}$,
$w_{\textsc{create}}=w_{\textsc{delete}}\in\{4,5,8\}$, and
$w_{\textsc{convert}}\in\{8,10,20\}$, keeping only the vectors that satisfy
\cref{eq:order}. Because no \textsc{convert} operation occurs on CubiCasa5K, the
$w_{\textsc{convert}}$ axis has no effect there and the sweep reduces to $21$
distinct weightings. The ranking fusion $<$ room-centric $<$ wall-first holds at
every one of them. The wall-first deficit never falls below $17.9\%$
relative to the room-centric system. The fusion lead over the
room-centric system is the one that varies, from $15\%$ to $1.7\%$ across the
sweep ($7.8\%$ at the default weights), smallest at the corner where a move
costs half a create ($2,3,4,8$), because the donor walls
that fusion adds are the ones that then need to be moved. The ordering is
invariant; the size of the fusion margin depends on the weights.

% ---------------------------------------------------------------------
\subsection{The ResPlan-FP benchmark}\label{sec:resplan}
% ---------------------------------------------------------------------
We convert the ResPlan corpus (data licensed
CC~BY~4.0) into a recognition benchmark that can be redistributed: $16{,}998$
plans (2 rejected for exceeding capacity), frozen splits of $14998/1000/1000$
(seed 42, id-hash fingerprints, full test-id list archived), a published render
protocol (median-thickness band, door arcs, window gaps, 256px), and evaluation
code. Converter watertightness on a probe set is $0.995$. The benchmark has
three baseline tracks (\cref{tab:resplan}):
\begin{itemize}
\item \textbf{Zero-shot} (models trained elsewhere): the fusion system reaches
  $0.688$; the wall-first (walls-only) arm alone reaches $0.640$ by sequence
  decoding and $0.621$ by graph readout (frozen CubiCasa5K thresholds), both with no
  failures, and the room-centric arm alone drops to $0.383$ with a $26.7\%$
  hard-failure rate (plans on which no room is emitted, scored as zero), a
  cross-domain failure of room generation. On the $733$ plans where the
  room-centric arm does decode, fusion adds $19.3$pp over it
  (\cref{tab:purity}).
\item \textbf{Finetuned}: a small in-domain finetune removes the failures
  ($26.7\%\to0\%$). The room-centric system reaches rooms/openings/watertight
  $0.986/0.989/0.986$ at wall F1 $0.928$, and the finetuned wall-first arm's
  sequence decode reaches wall F1 $0.968$. These clean-render scores serve to
  calibrate the ceiling and are not a claim about difficulty; the difficulty is
  in the zero-shot and real-scan settings.
\item \textbf{LLM zero-shot}: a frontier VLM (Gemini~3.1~Pro, July 2026,
  single-pass prompting on a 200-plan test subset, seed 42) scores $0.811$ at
  tolerance $0.05$ but $0.472$ at $0.015$. The topology is roughly right and
  the coordinates are coarse, which is the known VLM failure mode.
\end{itemize}

\begin{table}[t]
\centering
\caption{ResPlan-FP baseline tracks (wall F1 unless noted). The zero-shot
fusion row keeps the strictly zero-shot protocol (frozen-threshold readout
donor); fusion with the threshold-calibrated donor is an expected further
gain and is not listed as a track.}
\label{tab:resplan}
\small
\begin{tabular}{lcc}
\toprule
track & score & note \\
\midrule
zero-shot, fusion system      & 0.688 & readout fallback on failures \\
zero-shot, wall-first seq-decode alone & 0.640 & no failures \\
zero-shot, wall-first readout alone & 0.621 & frozen thr.; 0.677 calibrated \\
zero-shot, room-centric alone & 0.383 & 26.7\% hard failures \\
finetuned, room-centric       & 0.928 & rooms/open/wt $0.986/0.989/0.986$ \\
finetuned, wall-first seq-decode & 0.968 & --- \\
LLM zero-shot (Gemini 3.1 Pro) & 0.811 / 0.472 & @.05 / @.015; 200-plan subset \\
\bottomrule
\end{tabular}
\end{table}

% ---------------------------------------------------------------------
\subsection{Ablations}
% ---------------------------------------------------------------------
A matched pair of models trained from scratch, differing only in the presence
of the dense heads (synthetic validation set, $n{=}30$, snap $0.03$), shows
their effect and their cost. Adding the dense heads moves $r_{\mathrm{val}}$
from $0.03$ to $0.43$ and rooms by $+0.15$, at a cost of $0.024$ wall F1 and
$+9.6$ edit cost (move operations caused by snap displacement). This is the
trade the recipe makes to obtain watertightness. The set is small, so we
report the direction of the effect and not an interval.

% =====================================================================
\section{Limitations}\label{sec:limits}
% =====================================================================
The controlled study compares systems (native wall emission against room
emission followed by reconciliation), not representations in isolation, and
our claims are stated at the system level. Edit-cost magnitudes are set by
hand; the ordering holds across a weight sweep, but calibration remains
future work. The two systems of the controlled study are not equally mature: the
room-centric arm continues a published system from its released checkpoint,
while the wall-first arm is ours and its room and opening outputs are
under-tuned, so the study bounds what room emission with reconciliation loses
on wall structure and says nothing about rooms or openings. No arm varies
training supervision alone, so the earlier room-centric failures are
attributed to evaluation convention, reconciliation, and domain shift jointly.
Neither ink density nor plan size explains the cross-domain readout
reversal; it appears only where the decoder's training covered the render
style ($5$--$8$pp there), while under full domain shift the calibrated
readout stays ahead (\cref{sec:emitdetect}); the mechanism, presumably the
decoder's learned style prior, is not further tested. The conditioning
result in \cref{sec:cond} rests on ground-truth-content controls applied at test
time. The room-outline arm was trained on jittered ground-truth rooms and the
other two arms on a $60/20/20$ mix of cross-arm drafts, corrupted ground-truth
walls, and empty channels; no arm was trained on clean ground-truth content
alone, so we cannot exclude that corruption during training teaches the model
to discount the channel, although the room-outline arm, whose training content
was always ground-truth-derived, shows the same null result. We do not
report on the Structured3D official test split, because our pretraining overlaps
it, nor a trained official-CubiCasa leaderboard row, because our rooms are the
cited stage-1 output; CC~BY-NC-SA data is confined to the research path.
ResPlan-FP splits are frozen by plan id without near-duplicate removal;
ResPlan~v2 reports about 6.9\% near-duplicate plans across the corpus, so the
finetuned-track ceilings may be slightly inflated by near-duplicates across
splits. A trained junction-graph arm, which would turn the untrained readout
into a learned edge classifier with junction supervision as the primary
objective, is the natural next step, subject to a novelty check against HEAT,
Raster-to-Graph, and LETR in the stylized-raster, open-boundary, edit-cost
setting. ResPlan-FP so far carries only our own systems and one VLM as
baselines; third-party methods (HEAT, RoomFormer, Raster-to-Graph) have not
been run on it.

% =====================================================================
\section{Conclusion}
% =====================================================================
For rasterized floorplan vectorization, the choice of output representation is
less decisive than is often assumed. With matched data and recipe, room-centric
and wall-first systems reach almost the same wall quality; the remaining
differences mainly reflect (i) the representational envelope (what each format
can encode) and (ii) the evaluation convention and the post-processing step
(what a system is scored against, and what reconciliation or snapping
removes).

The factors that substantially affect performance are largely independent of
representation. Cross-system complementarity yields no gain as an input
prior, and costs 2--3 points when the draft contains errors, but delivers
$7$ wall-F1 points as output-level fusion. For a fixed trained network, reading
structure from dense detections outperforms autoregressive emission on large
real scans, where coordinate-by-coordinate emission is a measurable penalty,
while on small plans, and on clean renders whose style its training covered,
the decoder wins, for reasons that neither ink density nor plan size
explains. We release the edit-cost metric and the redistribution-clean
ResPlan-FP benchmark alongside these findings.

\appendix
% =====================================================================
\section{Edit-cost metric: procedure and breakdown}\label{app:edit}
% =====================================================================
\paragraph{Procedure.} Given a predicted plan $P$, a ground truth $G$, and a
matching tolerance $t$ (we use $t=0.05$ of the plan frame), the script
$s(P,G)$ is constructed as follows.
\begin{enumerate}
\item \textbf{Wall matching.} Predicted and ground-truth walls are matched by
  Hungarian assignment on the wall distance used by fpeval, with a move radius
  of $4t$ (one fifth of the frame at $t=0.05$). A predicted wall within this
  radius of a ground-truth wall is treated as that wall displaced, not as a
  delete followed by a create. The radius is deliberately generous; a tighter
  radius converts moves into delete/create pairs and raises all costs.
\item \textbf{Matched pairs.} Each pair contributes a \textsc{move} with
  magnitude equal to the matched distance in units of $t$, and a \textsc{type}
  operation if the wall types differ.
\item \textbf{Unmatched ground-truth walls} contribute one \textsc{create}
  each.
\item \textbf{Unmatched predicted walls.} If the wall's midpoint lies within
  the move radius of a ground-truth room-polygon edge that no ground-truth
  wall covers (an open boundary), it is a phantom wall over an open edge and
  contributes one \textsc{convert}; otherwise it contributes one
  \textsc{delete}. The second copy of a double wall therefore costs one
  \textsc{delete}. The reverse case, a missing wall on a real boundary, is
  charged as a \textsc{create} and not as a conversion.
\item \textbf{Openings} are matched by center distance within the same move
  radius, separately for doors and windows; matched pairs contribute an
  \textsc{open-move} (weight $0.5$ per unit of $t$), unmatched ground-truth
  openings an \textsc{open-create} ($4$), and unmatched predicted openings an
  \textsc{open-delete} ($4$).
\end{enumerate}
The reported edit cost is the per-plan total averaged over the test set. Wall
type is one of exterior, interior, partition, or shelter, as annotated in skv4.

\paragraph{Breakdown.} \Cref{tab:editops} gives the mean number of operations
per plan for the three systems on the CubiCasa5K official test ($n{=}287$, the
same roster as \cref{tab:fusion}), at the default weights.

\begin{table}[h]
\centering
\caption{Mean edit operations per plan, CubiCasa5K official test ($n{=}287$),
default weights. Rows are operation counts; the total is the weighted cost of
\cref{eq:edit}, in which a move is weighted by its displacement, so the total
is not a weighted sum of the rows. The wall-first total here ($110.2$) differs from
\cref{tab:main} ($113.9$, $n{=}296$) because of the roster; the room-centric
total is unchanged because \cref{tab:main} already scores it on these $287$
plans.}
\label{tab:editops}
\small
\begin{tabular}{lccc}
\toprule
operation & r2s\_mix & wf\_mix & fusion \\
\midrule
wall move            & 21.7 & 23.3 & 25.3 \\
wall type change     & 8.5  & 2.5  & 10.3 \\
wall create          & 7.5  & 6.0  & 4.0 \\
wall delete          & 1.8  & 5.1  & 3.1 \\
open-to-phantom convert & 0.0 & 0.0 & 0.0 \\
opening move         & 13.6 & 10.3 & 13.6 \\
opening create       & 1.3  & 4.6  & 1.3 \\
opening delete       & 1.3  & 4.0  & 1.3 \\
\midrule
total cost           & 79.3 & 110.2 & 73.1 \\
\bottomrule
\end{tabular}
\end{table}

% =====================================================================
\section{Synthetic data generator}\label{app:synth}
% =====================================================================
The synthetic arm of every \texttt{mix} and \texttt{synth-only} model is
produced by a procedural generator that takes a random seed and returns a
rendered raster plan, its vector ground truth in the fpeval format aligned to
the pixels, and a millimeter-space scene description from which the same plan
can be re-rendered in another style. Generation is deterministic given the
seed. The synthetic pool used here (\texttt{wf\_mix10k}) has 10{,}000
samples across all style families, all of which pass the fpeval
watertightness check; the mix training regime draws $3{,}327$ of them
(\cref{app:train}); the frozen evaluation tiers (\texttt{synth\_clean},
\texttt{synth\_marketing}, \texttt{synth\_brochure\_deg},
\texttt{synth\_openplan}) have 30 test plans each, generated from reserved
seeds that are excluded from training.

\paragraph{Unit statistics fitted to real listings.}
The generator starts from a unit description (bedroom and bathroom count,
target area, and which optional spaces are present). Rather than choosing
these uniformly, we fit them to real residential listings: 29{,}300 listing
titles from a Singapore property portal were parsed, 18{,}203 of which yielded
a bedroom/bathroom count. The joint distribution is given in
\cref{tab:synthmeta}. Target area is drawn from a triangular distribution over
a per-bedroom-count plausibility band (the bands are corpus-sanity ranges, not
regulation), and optional spaces are drawn with the prevalences in the same
table; the conditional ones (study, walk-in wardrobe) depend on bedroom
count. Only these aggregate statistics are used and released; no listing text
or image enters the generator or the paper. The fitted unit distribution is
what makes the synthetic corpus resemble a real housing stock at the level of
unit size and program. It is not a claim that the synthetic drawings resemble
any particular real drawing style; the evaluation sets used in this paper
(CubiCasa5K, ResPlan) come from other markets.

\begin{table}[h]
\centering
\caption{Unit statistics used by the generator. Left: bedroom/bathroom joint
distribution fitted to 18{,}203 parsed listings. Right: area bands
($\mathrm{m}^2$) per bedroom count and prevalence of optional spaces.}
\label{tab:synthmeta}
\small
\begin{tabular}{lr@{\hspace{2em}}lr}
\toprule
bed/bath & share & parameter & value \\
\midrule
0/1 (studio) & 0.5\%  & area band, 1 bed & 30--80 \\
1/1          & 15.2\% & area band, 2 bed & 50--130 \\
2/2          & 28.4\% & area band, 3 bed & 75--200 \\
3/2          & 36.0\% & area band, 4 bed & 110--350 \\
4/3          & 16.8\% & balcony & 0.70 \\
5/4          & 2.8\%  & service yard & 0.60 ($\ge$2 bed), 0.25 otherwise \\
             &        & household shelter & 0.75 \\
             &        & A/C ledge & 0.85 \\
             &        & study & 0.20 ($\ge$2 bed) \\
             &        & walk-in wardrobe & 0.50 ($\ge$3 bed) \\
             &        & store & 0.30 \\
\bottomrule
\end{tabular}
\end{table}

\paragraph{Layout.}
The unit description is expanded into a room program tree: a public zone
(foyer, living/dining, kitchen with its service cluster) and a private zone
(a corridor with the master suite, further bedrooms, bathrooms, and optional
study). The tree fixes the door topology in advance, so every room is
reachable by construction, and an area budget drops optional rooms a small
unit cannot afford. The envelope is a rectangle from area and aspect ratio
(1.0--2.2) on a 10\,mm grid, optionally with a corner notch (L-shaped plan,
$p\approx0.2$) or a corner chamfer (a non-Manhattan diagonal,
$p\approx0.22$). Rooms are realized by recursive guillotine splits along the
program tree, with each strip clamped to its child's minimum width; a corridor
strip of 0.95--1.35\,m is carved along the public-facing edge so that every
private room touches it. Room polygons share edges by construction. The wall
graph is obtained by noding all boundaries, splitting at junctions, and
classifying each segment as exterior (200--300\,mm), interior (100--150),
partition (75--100), or shelter (250--400 on all sides); collinear runs with
the same left room, right room, and class are merged. Each room keeps an
ordered cycle of wall IDs, which is what makes the ground truth watertight by
construction. With the open-plan option, shared boundaries between kitchen and
living/dining are left without a wall, so the room cycle carries an open edge;
this is the source of the \texttt{synth\_openplan} tier.

\paragraph{Openings and furniture.}
An entrance door is placed on an exterior wall of the foyer or living room;
interior doors follow the access tree with per-kind widths (entrance
900--1100\,mm, interior 750--950, shelter 700--800, sliding 1400--2600 to
balconies and yards). Windows are placed only on exterior walls of habitable
rooms. Doors swing into the room being entered, and a breadth-first search
over the door graph rejects any layout with an unreachable room. Furniture is
placed by rejection sampling against a per-room rule table (bed and wardrobe
in bedrooms, toilet, basin and shower in bathrooms, counter run, stove, sink
and fridge in kitchens, and so on), checked against free wall regions, placed
fixtures, and door-swing clearances. Room labels are drawn from a vocabulary of
printed plan labels with area strings; with $p\approx0.35$ one label is
deliberately shifted onto a wall band by a sampled overlap fraction, a
controlled occlusion condition recorded in the manifest.

\paragraph{Rendering and degradation.}
Plans are rendered in millimeters under a single affine transform in one of
four style families: clean CAD, marketing (floor fills, furniture-rich),
brochure (tinted paper, project title), and hand-drawn (stroke wobble only; the
geometry is unchanged). Wall bands are the centerline buffered by half the
thickness with opening gaps subtracted; door swings, sliding pairs, and one-,
two-, or three-line windows are drawn as symbols; dimension chains and a scale
bar carry true metric values. Degradation has two stages. The first is a
similarity transform (rotation $\le3^\circ$, scale 0.85--1.05, translation)
that is applied to the image and tracked exactly, so the ground truth is
transformed with it. The second is appearance-only: photocopy, ink bleed,
bleed-through, low-ink lines, lighting gradients, dirty rollers, dithering,
shadows, stains, blur, salt-and-pepper noise, line breakup, JPEG compression
(quality 40--92), down-up-scaling, a tiled watermark, and vignetting, sampled by
severity (none 0.30, light 0.30, medium 0.25, heavy 0.15). No second-stage
operation may displace pixels; a registration test enforces alignment within
1.5\,px through the full pipeline. The total image-to-ground-truth transform is
the product of the normalizer, the similarity, and the render affine, so
coordinates and lengths in the ground truth are exact.

\paragraph{Quality gate.}
Every sample passes through fpeval's verification before it is written; a
sample that fails watertightness or the reachability check is rejected and
regenerated from the next attempt of the same seed. The manifest records the
seed, unit description, style, degradation operations, transform, and
verification result for each sample.

% =====================================================================
\section{Evaluation definitions and the reconciliation step}\label{app:fpeval}
% =====================================================================
\paragraph{Coordinate frame.} Every plan, predicted or ground truth, is
expressed in a frame whose longer edge is $1024$ units. A tolerance $t$ is a
fraction of this frame; $t=0.05$ is $51$ units, which on a $10$\,m plan is
about $0.5$\,m, and $t=0.015$ is about $15$\,cm.

\paragraph{Wall matching.} Two walls are candidates for a match if the angle
between them is below $15^\circ$, the mean distance between seven points
sampled along one wall and the other wall's segment (averaged in both
directions) is at most $t$, and their projections onto the longer wall's axis
overlap by at least $30\%$ of the shorter wall's length. The cost of a
candidate pair is that mean distance; non-candidates have infinite cost.
Walls are assigned one-to-one by Hungarian matching, and a pair counts as
matched if its cost is at most $t$. Precision, recall, and F1 follow from the
matched count. This is the same matcher used by the edit-cost metric, there
with the move radius $4t$ in place of $t$.

\paragraph{Openings.} Door and window centers are matched within $2t$,
separately by family; the wider tolerance reflects that opening centers are
less precisely defined than wall endpoints.

\paragraph{Rooms.} Room polygons are matched one-to-one at IoU$\ge0.5$. For
the wall-first system, which emits rooms as cycles of wall IDs, the room
polygons are the faces of the planar graph formed by its walls. When a system
emits no rooms (the graph readout), rooms are again taken as the faces of
its wall graph.

\paragraph{Watertightness $r_{\mathrm{val}}$.} A plan is watertight if its
room polygons partition the interior: each polygon is simple, polygons do not
overlap beyond a small tolerance, and together they tile the region enclosed
by the exterior. A room boundary that has no wall along it (an open edge) is
allowed, so open-plan layouts can be watertight. $r_{\mathrm{val}}$ is the
fraction of plans that pass.

\paragraph{Ink gate.} A candidate wall's ink coverage is computed on the
$256$\,px input binarized at gray value $100$: the segment is sampled at one
point per $4$\,px (at least $5$ points), and a sample counts as covered if a
dark pixel lies within $\pm3$\,px of it perpendicular to the segment, which
admits hollow and double-line wall styles whose centerline carries no ink.
Coverage is the covered fraction; the gate keeps the wall if coverage
$\ge0.5$. The same binarization defines the dark-pixel fraction of
\cref{sec:emitdetect}.

\paragraph{Double walls.} Two predicted walls are counted as a double when
they are nearly parallel (the sine of the angle between them is at most
$0.2$), their perpendicular separation is at most $28$ frame units (about
$2.7\%$ of the frame), and they overlap along their common direction by more
than $40\%$ of the shorter wall. The same criterion decides whether a donor
wall overlaps a base wall during fusion (\cref{sec:fusion}).

\paragraph{Reconciliation step for the room-centric system.} Raster2Seq emits
each room as a separate polygon, so the two sides of a shared wall arrive as
two edges roughly one wall thickness apart (about $20$--$32$ frame units,
given its $32$-bin coordinate quantization at $256$\,px). The reconciliation
step converts these polygons into walls as follows. (1)~Coordinates are
un-padded and rescaled to the $1024$ frame. (2)~All polygon vertices are
welded by greedy centroid clustering with a tolerance of $0.022$ of the frame
($22$ units), chosen by a development-set sweep as the smallest
value that removes shared-edge duplicates without merging distinct walls.
(3)~Polygon edges are keyed by their welded endpoints and de-duplicated, so a
shared edge becomes one wall. (4)~Any edge that passes through a weld centroid
in its interior is split there (T-splitting), so that adjacent rooms which
subdivide a boundary differently still share walls. (5)~Edges shorter than
$0.002$ of the frame are dropped. Walls carry a constant thickness of $0.012$
of the frame, since room polygons carry no thickness information. Openings
and rooms are taken from the Raster2Seq output directly. The ``naive'' variant
in \cref{tab:postproc} skips steps (2)--(4).

% =====================================================================
\section{ResPlan-FP construction}\label{app:resplan}
% =====================================================================
\paragraph{Source.} ResPlan \citep{resplan2025} provides $17{,}000$
residential plans as vector geometry: wall bodies as unions of rectangles of a
per-plan wall depth, door, window, and front-door bodies that fill the gaps in
the wall band, and room polygons with labels. The data is licensed CC~BY~4.0
and the code MIT, which allows us to redistribute derived renders and labels.

\paragraph{Conversion to centerline walls.} Our ground truth represents walls
as centerline segments, so the wall bodies have to be converted. For each
plan, the wall bodies and the opening bodies are united into one continuous
band (the opening bodies fill the door and window gaps, so the band does not
break at openings). The band is rasterized at $4$ pixels per plan unit, closed
morphologically with a kernel of about half the wall depth, skeletonized, and
traced into a graph; the graph's paths are simplified with Douglas--Peucker
into straight segments. Segment endpoints are then welded at junctions,
snapped to the dominant axes with a tolerance of $0.35$ of the wall depth
(ResPlan plans are Manhattan), and degenerate segments are dropped. Openings
are projected onto the nearest centerline as intervals, keeping the door or
window family; front doors are doors. Rooms are kept as the source polygons
with their labels. The content bounding box is moved to the origin and the
longer edge scaled to the $1024$ frame.

\paragraph{Rendering.} Each plan is rendered at $256$\,px (longer edge) as a
clean line drawing: wall bands of the plan's median thickness, door swing
arcs, window gaps with jamb lines, no furniture and no text. The render and
the ground truth share the same transform.

\paragraph{Rejection and splits.} Plans whose converted sequence exceeds the
model's capacity ($512$ tokens or $80$ walls) are rejected; $2$ of $17{,}000$
were. The remaining $16{,}998$ are split $14{,}998/1{,}000/1{,}000$ by a seeded
shuffle (seed 42). Each split is stored with an id-hash fingerprint and the
full list of test ids is archived, so the benchmark can be checked against any
later copy. No near-duplicate removal is performed (\cref{sec:limits}). On a
$200$-plan probe the converted ground truth passes fpeval's watertightness
check at a rate of $0.995$; the failures are plans whose source polygons do
not tile exactly.

\paragraph{Tracks.} Zero-shot: models trained on CubiCasa5K and synthetic
data, evaluated on the $1{,}000$ test plans without any ResPlan training.
Finetuned: the same models continued on the $14{,}998$ training plans with
early stopping on validation (the wall-first arm for $108$ epochs).
LLM zero-shot: \cref{app:vlm}.

% =====================================================================
\section{Readout calibration}\label{app:readout}
% =====================================================================
The graph readout has three thresholds and one switch: the junction peak
threshold $\theta_J$, the centerline threshold $\theta_C$, the chord coverage
$\kappa$, and whether collinear edges are chained. All four are chosen on the
validation split of the domain being evaluated and then frozen before the
test split is run.

\paragraph{CubiCasa5K.} A $72$-configuration grid on the $318$ validation
plans, extended twice at the grid boundary until the best configuration lay in
the interior. Frozen configuration: $\theta_J=0.55$, $\theta_C=0.4$,
$\kappa=0.5$, no chaining; validation wall F1 $0.839$ at $t=0.05$ and
$0.810$ at $0.015$, $r_{\mathrm{val}}=0.959$. Chaining collinear edges costs
$22$pp of wall F1 at $t=0.05$ on validation, because the skv4 ground-truth
walls are split at junctions and chained walls then fail to match. The first
test invocation was accidentally run with chaining on (a flag slip visible in
the saved parameters) and was re-run once with the frozen configuration; no
parameter was changed after seeing test numbers.

\paragraph{ResPlan-FP.} Our first calibration on the ResPlan validation
split ($300$ plans) searched a grid that did not reach the CubiCasa5K operating
point and froze $\theta_J=0.3$, $\theta_C=0.35$, $\kappa=0.7$, no chaining
(test $0.881$ at $t=0.05$). Extending the grid until the optimum is interior
gives $\theta_J=0.65$, $\theta_C=0.3$, $\kappa=0.6$ (validation $0.913$;
test $0.915$ at $t=0.05$ and $0.896$ at $0.015$), and the frozen CubiCasa5K
configuration transfers almost as well (test $0.914$). A $0.717$ figure we
had previously reported for ``CubiCasa5K thresholds on ResPlan'' was a
configuration slip: it came from the harvest tool's command-line defaults
($\theta_J=0.25$, $\theta_C=0.3$, $\kappa=0.8$) with edge chaining
\emph{on}, not from the frozen CubiCasa5K configuration; chaining is what does the
damage there, as on CubiCasa5K. Consistent with thresholds transferring, the
junction heatmap responds no more weakly on the clean render: measured at
ground-truth wall endpoints, the finetuned network's junction peaks average
$0.86$ on ResPlan-FP ($95\%$ above $0.55$) against $0.80$ on CubiCasa5K
($81\%$) for the walls-only network, with centerline values at wall midpoints
of $0.66$ in both. With chaining off, the thresholds of a network used in its training domain
or finetuned in-domain transfer at a cost of at most $0.6$pp; for the
zero-shot network under domain shift the thresholds themselves are the
fragile part. Its in-domain optimum, calibrated on ResPlan validation with
grid extension until interior ($\theta_J=0.55$, $\theta_C=0.1$,
$\kappa=0.6$), scores $0.677$ on test against $0.621$ with the frozen CubiCasa5K
thresholds; the much lower centerline threshold compensates the weaker
heatmap response on the unseen domain (junction peaks at ground-truth
endpoints average $0.66$ zero-shot against $0.86$ after finetuning).

\paragraph{Open-plan tier.} The frozen tier has no validation split of its
own, and our first report used the ResPlan configuration unchanged ($0.777$). For the numbers in
\cref{tab:crossover} we generated an in-domain calibration set of $59$ plans
from a reserved seed (style and degradation as the tier, seed disjoint from
it) and ran the same grid, extended past its boundary until the optimum was
interior: $\theta_J=0.55$, $\theta_C=0.35$, $\kappa=0.5$, no chaining
(calibration-set wall F1 $0.851$). The frozen tier then scores $0.825$ at
$t=0.05$ and $0.753$ at $0.015$ ($r_{\mathrm{val}}=0.967$); the frozen
CubiCasa5K configuration transfers to $0.819$, so in-domain calibration is
worth $0.6$pp here, and a $0.561$ figure we had previously reported for
``CubiCasa5K thresholds'' on this tier was the same command-line-default slip
described above.
The mixed training set contains $3$ open-plan layouts among its $3{,}327$
synthetic plans (\cref{app:train}), so both readouts see the open boundary
essentially zero-shot.

% =====================================================================
\section{Training details}\label{app:train}
% =====================================================================
\paragraph{Wall-first network.} The network reuses the Raster2Seq backbone
(ResNet features, deformable-attention encoder) and causal autoregressive
decoder with anchor-based coordinate embedding, hidden size $256$, and
$32$-bin coordinate quantization per axis at $256$\,px input; the output
grammar (wall primitives, rooms as wall-ID cycles with open edges, openings
as along-wall intervals) is enforced by a type mask at each decoding step.
The dense branch predicts junction, centerline, and opening heatmaps at
$128\times128$ with sub-pixel offsets, and is trained jointly with the
sequence loss. At inference, sequence-emitted endpoints are snapped to the
nearest junction peak within a radius of $0.06$ of the frame. Training uses
batch $16$, learning rate $2\times10^{-4}$ with cosine decay and $3$ warmup
epochs, and early stopping on validation coordinate error with a patience of $15$ epochs, up to $500$ epochs. The
CubiCasa5K-trained arms, and their zero-shot rows, use the final checkpoint;
the best-validation checkpoint scores about $2$pp higher wall F1 on the
CubiCasa5K test and is not used for them. The ResPlan-finetuned arm uses its
early-stopped best-validation checkpoint, for the decoder and the dense heads
alike, so every row of \cref{tab:crossover} compares two readouts of one and
the same checkpoint. All sequence-decode numbers use the recipe's snap radius
of $0.06$ (the decoder's code default is $0.03$). All arms are initialized from a model
pretrained on Structured3D layouts rendered through the synthetic renderer;
this pretraining overlaps the Structured3D official test split, so
we do not report on it.

\paragraph{Room-centric arms.} Raster2Seq is trained with its published
recipe (polygon-sequence objective with per-token semantic loss, $12$
semantic classes, $512$-token sequences, $32$ bins, EMA weights for
evaluation) on the same data regimes, $500$ epochs at batch $56$, starting
from the published checkpoint.

\paragraph{Data regimes.} \texttt{cc5k}: the $3{,}328$ official training
plans with skv4 annotations. \texttt{cc5k+synth} (``mix''): the same plans
and $3{,}327$ synthetic plans drawn (seed $42$) from the $10{,}000$-plan
pool, one to one; the subset contains $3$ open-plan layouts. The
room-centric mix arm was trained on a merged COCO export of the same two
sources; its training log runs $118$ iterations per epoch at batch $56$
($\approx6{,}608$ samples $\approx3{,}328+3{,}327$), confirming the same
one-to-one composition.
\texttt{synth-only}: the synthetic plans alone. The conditioning arms of
\cref{sec:cond} and the walls-only arm used as the graph-readout network are
trained on the \texttt{mix} regime.

\paragraph{CubiCasa5K annotations (skv4).} The official CubiCasa5K
annotations give rooms and walls as SVG polygons. Two conversion problems
were found and corrected. First, room polygons converted as the inner faces
of the wall polygons did not close into wall cycles ($44\%$ of room cycles
closed; $99.4\%$ of plans failed watertightness), which gave the sequence
model incoherent room supervision; rooms are now built as the faces of the
welded wall graph and openings are projected onto the wall centerlines
(skv3). Second, wall centerlines derived from the skeleton carried spurs and
segments without ink under them; an ink-coverage gate and spur pruning
remove these (skv4). The official train, validation, and test id lists are
kept unchanged ($3328/318/296$ of the plans that convert). Re-evaluating a
fixed checkpoint against the corrected ground truth moved wall F1 at $t=0.05$
from $0.74$ to $0.825$ and opening F1 from $0.29$ to $0.66$; this is the
evaluation-convention factor (i) of \cref{sec:refute}. It is a statement
about scoring, not about training: the same checkpoint scored differently,
and no arm trained under the old annotations was compared with one trained
under the corrected ones. The converter and the corrected
annotations are released with the code.

% =====================================================================
\section{LLM zero-shot protocol}\label{app:vlm}
% =====================================================================
The frontier-VLM row of \cref{tab:resplan} uses Gemini~3.1~Pro through an
API gateway, in July 2026, with temperature $0$, a $16{,}000$-token output
budget, and one call per plan with no retries or self-correction. The image
is the $256$\,px benchmark render upscaled to a longer edge of $1024$\,px
and sent as PNG. The prompt is:

\begin{quote}\small
You are an expert at reading architectural floor plans. This image is a
single residential unit floor plan. First, briefly reason: list the rooms
you see and describe the building outline and interior wall layout. THEN
output the final vectorization as a single JSON object and nothing after
it, with this schema:
\begin{verbatim}
{"walls":[{"id":"wall_N","start":[x,y],"end":[x,y],
           "thickness":T,"curvature":0,
           "openings":[{"type":"door|window",
                        "center":[x,y],"width":W}]}],
 "rooms":[{"label":"room_type","walls":["wall_N"]}]}
\end{verbatim}
Trace EVERY structural wall (building outline + all interior partitions);
ignore furniture, fixtures, text and colour fills. Include every door and
window as an opening on its wall. Coordinates are normalized so the longer
image edge = 1024.
\end{quote}

The JSON object is extracted from the end of the response and parsed into the
fpeval plan format; a response with no parseable object scores zero on that
plan. The $200$ test plans are a seeded random subset (seed 42) of the
$1{,}000$-plan test split, and their ids are archived with the benchmark. On
this subset the model scores wall F1 $0.811$ at $t=0.05$ and $0.472$ at
$0.015$, rooms $0.708$, openings $0.752$, $r_{\mathrm{val}}=0.620$, at a
total cost of about US\$6. Gemini was chosen as the sole representative
after an earlier comparison with GPT-5.1 and Qwen-VL on the CubiCasa5K
baseline matrix, where it was the strongest of the three.

% =====================================================================
\section{Fusion control rosters}\label{app:rosters}
% =====================================================================
The fusion steps in \cref{tab:fusion} are computed on the $287$ of $296$
test plans on which both the room-centric and the wall-first model produce a
decodable output. The ensembling controls of \cref{sec:fusion} require a
third model (the synthetic-only room-centric arm) and are computed on the
$283$ plans on which all sources decode. On that roster the base scores are
$0.786$ (r2s\_mix) and $0.793$ (wf\_mix) at $t=0.05$, cross-representation
fusion reaches $0.855$ (a gain of $6.9$pp over r2s\_mix) and the homogeneous
r2s ensemble $0.844$ ($5.8$pp). The within-table comparison is the point;
absolute values differ from the $287$-plan ladder by about $0.005$ because
the four dropped plans are difficult ones. The same-representation ensemble
control ($0.794$) and the $5.9$pp figure in the main text are on the
$287$-plan roster.

\bibliographystyle{unsrtnat}
\bibliography{references}

\end{document}